\pdfoutput=1

\documentclass[11pt]{article}

\usepackage[review]{acl}

\usepackage{times}
\usepackage{latexsym}

\usepackage[T1]{fontenc}

\usepackage[utf8]{inputenc}

\usepackage{microtype}

\usepackage{bbm}
\usepackage{amsmath,amssymb,amsfonts}
\usepackage{subfigure}
\usepackage{algorithmic}
\usepackage{graphicx}
\usepackage{textcomp}
\usepackage{xcolor}
\usepackage{tabularx}
\usepackage{hyperref}
\usepackage{multirow}
\usepackage{caption}
\title{Temperature-scaling surprisal estimates improve fit to human reading times -- but does it do so for the ``right reasons''?}
\author{Tong Liu$^{1,2}$, \ Iza Škrjanec$^{3}$, \ Vera Demberg$^{3,4}$ \\
$^1$LMU Munich \quad $^2$Munich Center for Machine Learning \quad $^3$Saarland University 
\\ 
$^4$Max Planck Institute for Informatics, Saarland Informatics Campus\\
        \texttt{tongliu.physics@gmail.com},
\texttt{\{skrjanec,vera\}@coli.uni-saarland.de} }

\begin{document}
\nolinenumbers
\newtheorem{theorem}{Theorem}
\newtheorem{lemma}[theorem]{Lemma}

{\makeatletter\acl@finalcopytrue
  \maketitle
}
\begin{abstract}

A wide body of evidence shows that human language processing difficulty is predicted by the information-theoretic measure \textit{surprisal}, a word’s negative log probability in context. However, it is still unclear how to best estimate these probabilities needed for predicting human processing difficulty -- while a long-standing belief held that models with lower perplexity would provide more accurate estimates of word predictability, and therefore lead to better reading time predictions, recent work has shown that for very large models, psycholinguistic predictive power decreases. One reason could be that language models might be more confident of their predictions than humans, because they have had exposure to several magnitudes more data.
%
In this paper, we test what effect temperature-scaling of large language model (LLM) predictions has on surprisal estimates and their predictive power of reading times of English texts. Firstly, we show that calibration of large language models typically improves with model size, i.e.~poorer calibration cannot account for poorer fit to reading times. Secondly, we find that temperature-scaling probabilities lead to a systematically better fit to reading times (up to 89\% improvement in delta log likelihood), across several reading time corpora. Finally, we show that this improvement in fit is chiefly driven by words that are composed of multiple subword tokens.\footnote{Code available at:~\url{https://github.com/TongLiu-github/TemperatureScaling4RTs}.}
\end{abstract}

\section{Introduction}
\label{sec:introduction}

In psycholinguistics, a key finding is that words with higher surprisal (= negative log probability of the word in context) require more time for processing~\citep{hale2001probabilistic, levy2008expectation}. 
Numerous studies provided experimental evidence supporting this theory, demonstrating that surprisal is a powerful predictive measure of processing complexity~\citep[e.g.,][]{demberg2008data, wilcox2020predictive, wilcox2023testing, shain2022large}, and that the relationship between surprisal and reading times (RTs) seems to be linear~\citep{smith2013effect, wilcox2020predictive, shain2022large}.

However, prior work implicitly made the assumption that human predictability estimates would be similar to the actual probability of a word occurring in a given context, and that therefore, surprisal values estimated from models that achieve lower perplexities should also approximate human processing difficulty better \citep{goodkind2018predictive, merkx-frank-2021-human}. 

Recent research has however found that this is not true -- surprisal values from very large LLMs provide in fact a very poor fit to reading times. \citet{oh2023does} hypothesize that this might be due to LLMs being ``too confident'' in their estimates of rare named entities compared to humans, thanks to their manifold larger exposure to data and greater memory capacity compared to humans.
Furthermore, work on NLP applications like question answering has reported that probability estimates from pretrained language models are often overconfident, i.e.~they are higher than the ground truth probability~\citep{si2022re, kumar2022answer}. These findings hence beg the question whether current LLMs are well-calibrated with respect to ``objective'' word occurrence probabilities. Relatedly, we ask whether LLM probability estimates are overconfident compared to human estimates (as observed in reading times). 

%


One approach to address calibration problems is to use \textit{temperature scaling}, as done e.g., in vision tasks \citep{guo2017calibration, hendrycks2019using}. Temperature-scaling with a temperature $T>1$ has the effect that the probability distribution is flattened such that it becomes more similar to a uniform distribution. Temperature-scaling hence incorporates uncertainty into the probability estimates from LLMs. 


We note that the idea to work with flattened distributions instead of the original probability distributions from LLMs is also related to contextual R\'enyi Entropy as discussed by \citet{pimentel-etal-2023-effect}, as well as the super/sub-linear surprisal effect by \citet{shain2022large,hoover2023plausibility}. However, rather than merely adjust the power of surprisal in super/sub-logarithmic patterns or the power of probability in R\'enyi entropy, our work represents 
a distinct branch of study (i.e., probability calibration) in machine learning: shaping the probability distribution itself through shaping the logits before softmax. 
We also discuss the motivation for why a slightly flattened distribution may be more suitable, and whether this change in distribution is applied when calculating surprisal vs.~when calculating entropy. 

Our experimental results show that scaling probabilities can largely improve the fit to reading times in all 12 settings (3 corpora $\times$ 4 neural LMs). 
Our contributions are summarized as follows: 
(1) We propose temperature-scaled surprisal, where surprisal is calculated from temperature-scaled probabilities. 
(2) We demonstrate that temperature-scaling with temperature T$\approx$2.5 improves predictability of human reading times  of English texts compared to T=1.
(3) We identify linguistic phenomena that correlate with  the benefit of temperature-scaled surprisal by analyzing residual errors from regression models. 

\section{Predictive Power for Reading Times}
\label{sec:background}
In psycholinguistics, RTs on a word are believed to correlate with its processing difficulty. 
RTs can be gathered using different paradigms, including eye-tracking while reading text on a screen~\citep{rayner1998eye}, self-paced reading~\citep{aaronson1976performance, mitchell1978effects} and the Maze task~\citep{forster2009maze}.

The most common procedure for predicting words' RT is first to select a set of predictor variables thought to impact RTs $\mathbf{v} = [v^{(1)}, ..., v^{(d)}]^{\top} \in \mathbb{R}^{d}$, which include, e.g., the length of a word $w_{t}$, $|w_{t}|$, the frequency of a word $\mathrm{freq}(w_{t})$. 
Let $f_{\phi}: \mathbb{R}^{d} \rightarrow \mathbb{R}$ be a regression model parametrized by $\phi$ used to fit these predictors for the prediction of human RTs $rt$: $rt(w_{t}|\boldsymbol{w}_{<t}) \sim f_{\phi}(\mathbf{v})$, given the previous context $\boldsymbol{w}_{<t}$. 
The performance of $f_{\phi}$ is quantified by its log-likelihood, with a higher log-likelihood indicating a better psychometric predictive power for human RTs~\citep{frank2011insensitivity, fossum2012sequential}. 

Besides the word length $|w_{t}|$ and word frequency $\mathrm{freq}(w_{t})$, a word's surprisal (i.e., its negative log-probability in context)~\cite{hale2001probabilistic, levy2008expectation} has been shown to be predictive of RTs~\citep{demberg2008data, goodkind2018predictive, wilcox2020predictive, shain2022large}.  

\section{Methods}
\label{sec:methods}

In this section, we delve into key aspects of information-theoretic measures in language comprehension. 
We start with surprisal, a method connecting processing difficulty to word predictability. 
As word predictability is empirically estimated by LLMs, we introduce the notion of calibration errors, metrics quantifying how good the estimation of word predictability is. 
Further, we lay out temperature-scaled surprisal, and the relation between varying temperature vs.~varying $\alpha$ in contextual R\'enyi entropy.

\subsection{Surprisal}
Starting from~\citet{shannon1948mathematical}, the information conveyed by a word $w_{t}$ has been quantified as the negative log probability of the word $w_{t}$ given its previous context $\boldsymbol{w}_{<t}$. In Surprisal Theory~\citep{hale2001probabilistic, levy2008expectation}, this quantity is called surprisal $s(w_t)$ and proposed to be predictive of the  word's processing difficulty, typically quantified as its RT. 
Surprisal values are typically estimated from language models $\hat{p}(w_{t}|\boldsymbol{w}_{<t})$. 
\begin{align}
s(w_{t}) &= -\mathrm{log}_{2}~{p(w_{t}|\boldsymbol{w}_{<t})},
\label{eq:surprisal}
\end{align}

\subsection{Calibration error}
\paragraph{Definitions}
Let $\mathcal{D} = \{(x_{i}, y_{i})\}_{i}^{N}$ be a data set where $x_{i} \in \mathcal{X}$ is an sample (i.e., context) and $y_{i} \in \mathcal{K} = [K]$ is a category label. 
Let $g_{\theta}$ and $\hat{\mathbf{z}}_{i} = g_{\theta}(x_i)$ denote a language model parametrized by $\theta$ and the output logit vector of sample $i$, respectively. 
The predicted class label $\hat{y}_i$ for sample $i$ is given by $\hat{y}_i =\mathop{\arg\max}_{k \in \mathcal{K}}g(x_i)_k$ and confidence for sample $i$ is given by $\hat{p}_i =\mathop{\max}_{k \in \mathcal{K}}g(x_i)_k$. 
A model is perfectly calibrated when the confidence $\hat{p}$ is equal to the frequency of correctness, i.e., $\mathbb{P}(\hat{y}_{i} = y_{i} | \hat{p}_{i} = p) = p$ holding for all $p \in [0, 1]$ and any sample $i$. Any difference between the left and right sides of the above equation indicates there exists a \textit{calibration error}. 

\paragraph{Expected calibration error (ECE) \citep{guo2017calibration}} ECE is the most popular calibration metric, which empirically approximates the calibration error by discretizing the probability interval into a fixed number of bins ($B_{m}$ with $m \in \{ 1, 2, ..., M\}$), and measures the gaps of averaged confidence and averaged accuracy in each bin $B_{m}$. 
\begin{equation}
\mathrm{ECE} = \frac{1}{N} \sum_{m=1}^{M} |\sum_{i \in B_{m}} \hat{p}_{i} - \sum_{i \in B_{m}}\mathbbm{1}[\hat{y}_{i} = y_{i}]|,
\label{eq:ece}
\end{equation}
where $\mathbbm{1}$ is the indicator function. However, it does not necessarily measure the actual-word probability, which is the probability required for calculating surprisal in Eq.~\ref{eq:surprisal}. 
It focuses only on the top-label probability for a given sample. 

\paragraph{Classwise-ECE (CECE) \citep{kumar2019verified, kull2019beyond}} In comparison, CECE measures probabilities of all classes. 
For each bin and every class $k$, it assesses the difference between the average confidence
of samples for class $k$ and the actual proportion of class $k$. If assuming all classes weigh equally, we have: 
\begin{align}
\begin{split}
\mathrm{C}&\mathrm{ECE} \\
&= \frac{1}{NK}\sum_{k=1}^{K}\sum_{m=1}^{M}  |\sum_{i \in B_{m}} \hat{p}_{i, k} - \sum_{i \in B_{m}}\mathbbm{1}[k = y_{i}]|,
\label{eq:cece}
\end{split}
\end{align}
where $\hat{p}_{i, k}$ is the predicted probability of sample $i$ for class $k$.  



\paragraph{Human-likeness calibration error (HCE)} 
We define the HCE as the Kullback-Leibler divergence (KL divergence) between predicted probability $\boldsymbol{\hat{p}}$ from a neural LM and actual probability $\boldsymbol{p}^{*}$ of \textit{human} language model. 
\begin{equation}
    \mathrm{HCE} = D_{KL}(\hat{\boldsymbol{p}} || \boldsymbol{p}^{*}).
\end{equation}
Empirically, since $\boldsymbol{p}^{*}$ is not directly observable, we approximate it by the estimates of a temperature-scaled model that best fits human reading times (as discussed later). 
We denote the approximated $\mathrm{HCE}$ using such a method as $\mathrm{HCE}_{\mathrm{TS}}$. 


\subsection{Temperature-scaled surprisal}
Temperature scaling~\citep{guo2017calibration} is a widely-used method to improve model calibration. Given the output logit vector $\hat{\mathbf{z}}_{i}$ for sample $i$, a single scalar $T > 0 $ is applied to rescale $\hat{\mathbf{z}}_{i}$ before the softmax activation:
\begin{align}
\hat{q}_i = \underset{k}{\max}\,\sigma_{SM} (\frac{\hat{\mathbf{z}}_{i}}{T})^{(k)},
\label{eq:TS}
\end{align}
where $\hat{q}_i$ is the calibrated confidence for sample $i$, and $\sigma_{SM}$ is the softmax function. 
Scaling by a scalar $T$ does not alter the ranking; hence, the predicted label $\hat{y}_{i}$ remains unchanged. 
As $T>1$, it ``softens'' the probability distribution (i.e., makes the distribution \textit{more uniform}), \textit{increasing uncertainty and entropy} of the probability distribution, while $T<1$ peaks the distribution. The parameter $T$ in research on calibration is optimized by minimizing the negative log-likelihood on the validation set. 
In our experiments of fit to human RTs, we manually tune this temperature with $T>1$.   

Temperature scaling has been successfully applied in several applications: In knowledge distillation~\citep{hinton2015distilling}, temperature scaling (with $T>1$) is used to ``soften'' the knowledge (i.e., probability distribution) provided by the teacher model; in text generation, temperature is used to shape the probability distribution to ease certain aspects of the problems of top-k sampling (e.g., choosing an appropriate $k$ value across varying contexts)~\citep{ficler2017controlling, fan2018hierarchical}. 
Temperature tuning inherently shifts the model's output in the generation's quality/diversity spectrum~\citep{caccia2018language}, with higher temperature decreasing the quality of generation while improving its diversity. 
This also aligns with our consideration of a possibility that human probability distributions might be flatter than the ones learned by language models and thus increasing the predictive diversity of surprisal provided by LLMs could potentially yield more human-like distributions.

Given Eq.~\ref{eq:TS}, temperature-scaled surprisal is: 
\begin{align}
s_{T}(w_{t}, T) =  -\mathrm{log}_{2}{(\sigma_{SM}(\hat{\mathbf{z}}_{w_{t}}/T)^{(k^{*})}}),
\label{eq:surprisal_TS}
\end{align}
where $\hat{\mathbf{z}}_{w_{t}}$ and $k^{*} = y_{w_{t}}$ denote the logit vector and the actual word $w_{t}$ class, respectively. 
For given $t \in (0, \infty)$, we simply denote $s_{T}(w_{t}, T = t)$ as $s_{T}|_{T = t}$. 
A temperature $T$ with its best performance of final fit to RTs is denoted as $T^{*}$.

\begin{figure}[]
\centering 
\hspace{-5mm}
\vspace{-5mm}
\subfigure{
\includegraphics[width=0.7\linewidth]{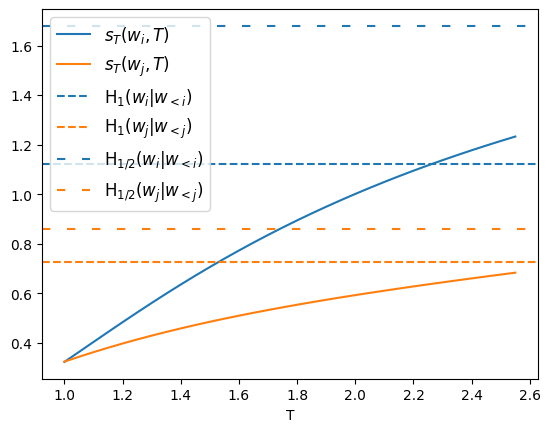}}
\hspace{-5mm}
\caption{Temperature-scaled surprisal $s_{T}(w_{t}, T)$ with corresponding $T \in [1, 2.5]$ for two random five-class probability distributions: $p_{i} = [0.8, 0.05, 0.05, 0.05, 0.05]$ and $p_{j} = [0.8, 0.2, 0, 0, 0]$. Dashed lines show Shannon entropy ($\mathrm{H}_{1}$). Loosely dashed lines show Rényi entropy with $\alpha = 1/2$ ($\mathrm{H}_{1/2}$). }
\label{fig:TS_surprisal}
\end{figure}

The extent to which a word's surprisal is affected by temperature scaling depends on the distribution and thus correlates with the entropy at word $w_t$. 
Consider an example of two five-class probability distributions  $\boldsymbol{p}_{i} = [0.8, 0.05, 0.05, 0.05, 0.05]$ and $\boldsymbol{p}_{j} = [0.8, 0.2, 0, 0, 0]$, for which the word indicated by the first position in the probability vector has identical surprisal in both $\boldsymbol{p}_i$ and $\boldsymbol{p}_j$.
Notably, $\boldsymbol{p}_{i}$ is more uniform and $\boldsymbol{p}_{j}$ is more peaked, resulting in distinct entropy characteristics: $\mathrm{H}(w_{i}|\boldsymbol{w}_{<i}) > \mathrm{H}(w_{j}|\boldsymbol{w}_{<j})$, where the entropy defined as the expectation of surprisal of current word $w_{t}$ over vocabulary, $\mathrm{H}(w_{t}|\boldsymbol{w_{<t}}) = \mathbb{E}_{w^{'} \sim p(\cdot| \boldsymbol{w}_{<t})} [s(w^{'})] = -{\sum_{w^{'} \in \overline{\mathcal{W}}}p( w^{'}| \boldsymbol{w}_{<t})} \log_{2}p(w^{'}| \boldsymbol{w}_{<t})$, where $\overline{\mathcal{W}} = \mathcal{W} \cup\{\mathrm{EOS}\}$ denotes the set of vocabulary $\mathcal{W}$ with EOS token. 
Fig.~\ref{fig:TS_surprisal} illustrates \textbf{a greater increase in surprisal} for a word with a more uniform distribution than with a more peaked distribution. 

This figure also anecdotally shows that the effect of applying temperature scaling with $T>1$ is similar to the effect of setting $\alpha <1$ in R\'enyi entropy. We will discuss the relationship between these parameters in more detail in Appendix~\ref{sec:appx_theorems}.

\section{Experimental setup}
\label{sec:experimental_setup}

\subsection{Datasets}
We conduct analyses on two self-paced reading corpora, the Natural Stories Corpus~\citep{futrell2018natural} and the Brown Corpus~\citep{smith2013effect}, as well as on the Dundee Corpus~\citep{kennedy2003dundee}, which contains the eye-movement record; our analyses in this paper focus on first-pass times\footnote{First pass times are calculated as the sum of all fixation durations from first entering to first leaving the word during the first pass, i.e., only those cases are counted where no words further advanced in the text have been fixated.} from the Dundee corpus.
We follow previous work with respect to the preprocessing steps for each corpus~\citep{kuribayashi2022context, shain2022large}. 
Appendix~\ref{sec:appx1} includes details about the preprocessing steps of each corpus.

\subsection{Language Models}
Recent observations showed that surprisal provided by LLMs with more parameters and lower perplexity is less predictive of self-paced reading times and eye-gaze durations~\citep{shain2022large, oh2023does}; across different experiments, \mbox{GPT-2} \citep{radford2019language} surprisals were found to predict human RTs the best. 
Therefore, we take four variants of pretrained GPT-2 (small, medium, large,  xl) as our language models in all experiments. 
Following prior work, we obtain the surprisal for words composed of more than one subword by summing up the surprisal estimates of the subwords.

\subsection{Metrics and evaluation}
We measure the predictive power of surprisal estimates from different language models, which is denoted as the log-likelihood difference per data point between a linear mixed-effects (LME) regression model using lme4 package~\citep{bates2015fitting} with a predictor of surprisal estimates (target model) and a model without surprisal (base model), following~\citet{goodkind2018predictive, wilcox2020predictive}. 
More specifically, the metric of delta log-likelihood is defined as:
\begin{align}
\Delta_{\mathrm{llh}} = \mathrm{llh}(f_{\phi}(\mathbf{v}^{tgt})) - \mathrm{llh}(f_{\phi}(\mathbf{v}^{base})),
\label{eq:PPP}
\end{align}
where $\mathbf{v}^{tgt}$ is \textit{target predictor variables} that include baseline predictor variables as well as predictor variables of our interest, such as surprisal or temperature-scaled surprisal. 
$\mathbf{v}^{base}$ is \textit{base predictor variables} only including baseline predictor variables. 
The greater the value of $\Delta_{\mathrm{llh}}$, the more valuable the additional surprisal estimates are for predicting human reading times. 

For the calibration error evaluation, we set the number of bins $M$ to $15$ for both ECE and CECE, aligning with prior literature, such as works by~\citet{guo2017calibration, kumar2019verified, rahimi2020intra}, to ensure consistency in addressing problems where comparable probability ranges are relevant. 
The calibration metrics (ECE and CECE) are evaluated separately on each of the reading time corpus $\mathcal{D}$.
For simplicity, our calibration evaluation is conducted at the token level. 
Given that many words have extremely low probabilities and thus are often grouped into a single bin, we also evaluate the calibration error \textit{under the log probability binning scheme}. 
For further descriptions regarding the metrics and evaluation, see Appendix~\ref{sec:appx_other_metric}.

\section{Results}
\label{sec:results}

\subsection{Calibration of LLMs}
Table~\ref{tab:llm_improvement} shows ECE and CECE in log binning scheme for GPT-2 models of different sizes. 
\textbf{LLMs are in general well calibrated on language modeling.} Besides, \textbf{LLM calibration improves with scale.} Larger LMs are better calibrated. 
This conclusion is consistent with calibration investigation evaluated in BIG-bench multiple-choice tasks in~\citet{srivastava2023beyond} as well as in several tasks including language modelling in~\citet{zhu-etal-2023-calibration}. 

\subsection{Main result: temperature-scaled surprisal improves human reading time prediction}
\label{sec:results1}
\begin{figure*}[t]
\centering 
\hspace{-5mm}
\subfigure{
\includegraphics[width=0.35\textwidth]{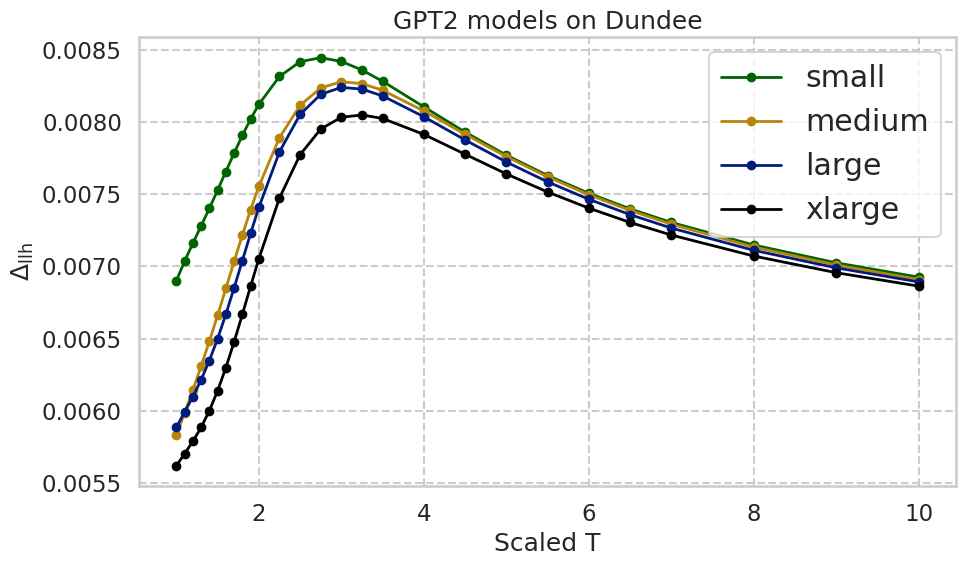}}
\hspace{-5mm}
\subfigure{
\includegraphics[width=0.35\textwidth]{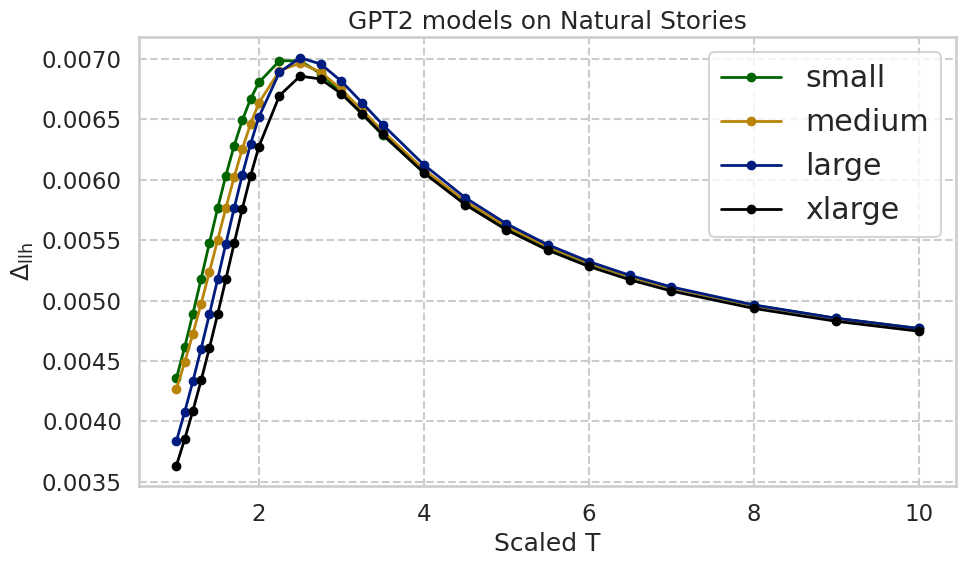}}
\hspace{-5mm}
\subfigure{
\includegraphics[width=0.35\textwidth]{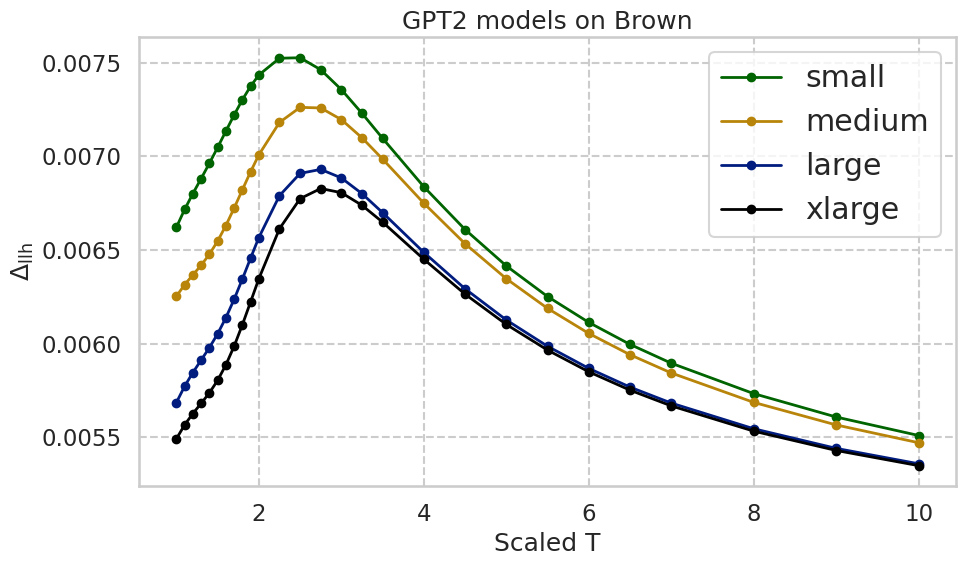}}
\hspace{-5mm}
\vspace{-3mm}
\caption{Relationship between $\Delta_{\mathrm{llh}}$ of GPT-2 models
and corresponding temperature. T is scaled from 1.0 to 10.}
\label{fig:main_result1}
\end{figure*}

We evaluate the predictive power of temperature-scaled surprisal. 
We scale $T$ in the range of $[1, 10]$ and measure $\Delta_{\mathrm{llh}}$, see Fig.~\ref{fig:main_result1}.
First, a confirmatory observation regarding the relationship between model size and predictive power: 
At $T = 1$, GPT-2 small exhibits the best predictive performance, and as the model size increases, $\Delta_{\mathrm{llh}}$ declines, which is consistent with previous studies~\citep{shain2022large, oh2022comparison, oh2023does}. 
Secondly, \textbf{scaling the surprisal with $\boldsymbol{T>1}$ can significantly improve the predictive power across all corpora and LLMs.}
With optimal $T^{*}$, on Dundee, Natural Stories, and Brown, the $\Delta_{\mathrm{llh}}$ improvement is 23-43\%, 60-89\%, and 14-24\%, respectively. 
We assess statistical significance of GPT-2 small in Appendix~\ref{sec:appx_significance}, where we report a result of p < 0.001 on three corpora.   
We also observe a consistent pattern: 
when increasing $T$, $\Delta_{\mathrm{llh}}$ first rises then declines; \textbf{the optimal value $\boldsymbol{T}^{*}$ falls within the range of (2, 3) (around 2.5) across all models and corpora} in our setting. 
At $T^{*}$, even though the impact of model size on final performance is not fully recovered, the disparity diminishes. 
Smaller models continue to outperform, but the extent of model sizes influencing performance is reduced.

Finally, \textbf{larger LMs typically have a larger human-likeness calibration error}, shown in Table~\ref{tab:llm_improvement}. 
Larger LMs also require a higher value of $\mathrm{T}$ to reach their best performance and have a greater increase by temperature-scaled surprisal.

\begin{table}[t]
\tabcolsep=2pt
\centering
\small
\begin{tabular}{ccccccr}
\hline
& &  $T^{*}$ & $\Delta_{\mathrm{llh}} +$ & $\mathrm{HCE}_{\mathrm{TS}}\downarrow$ & $\mathrm{ECE}_\mathrm{log}\downarrow$ & $\mathrm{CECE}_\mathrm{log}\downarrow$
\\ \hline
\multirow{4}{*}{Dundee} & \multicolumn{1}{l|}{s} & 2.75 & 22.5 & 3.11 & 1.59 & 4.07E-03
\\
& \multicolumn{1}{l|}{m} & 3.0 & 42.0 & 3.61 & 1.74 & 4.13E-03
\\
& \multicolumn{1}{l|}{l} & 3.0 & 39.9 & 3.82 & 1.55 & 3.99E-03
\\
& \multicolumn{1}{l|}{xl} & 3.25 & 43.2 & 4.13 & 1.29 & 3.84E-03
\\ \hline
\multirow{4}{*}{NS} & \multicolumn{1}{l|}{s}  & 2.5 & 60.3 & 3.31 & 1.91 & 1.53E-02
\\
& \multicolumn{1}{l|}{m} & 2.5 & 63.0 & 3.50 & 1.80 & 1.50E-02 \\
& \multicolumn{1}{l|}{l} & 2.5 & 82.6 & 3.97 & 1.70 & 1.40E-02
\\
& \multicolumn{1}{l|}{xl} & 2.5 & 89.0 & 4.07 & 1.56 & 1.35E-02
\\ \hline
\multirow{4}{*}{Brown} & \multicolumn{1}{l|}{s} & 2.5 & 13.7 & 3.10 & 1.69 & 1.53E-02
\\
& \multicolumn{1}{l|}{m} & 2.5 & 16.2 & 3.29 & 2.27 & 1.51E-02
\\
& \multicolumn{1}{l|}{l} & 2.75 &21.8 & 4.18 & 1.58 & 1.44E-02
\\
& \multicolumn{1}{l|}{xl} & 2.75 & 24.4 & 4.29 & 1.56 & 1.38E-02
\\ \hline
\end{tabular}
\caption{Optimal $T^{*}$, $\Delta_{\mathrm{llh}}$ improvement (\%) ($\Delta_{\mathrm{llh}} + = (\Delta_{\mathrm{llh}}(T=T^{*}) - \Delta_{\mathrm{llh}}(T=1))/\Delta_{\mathrm{llh}}(T=1)$), and 
calibration errors ($\mathrm{HCE}_{\mathrm{TS}}$, \% ECE and \%
CECE) for GPT2s on Dundee, Natural Stories (NS) and Brown. $\Delta_{\mathrm{llh}}$ values are multiplied by 1000. ECE and CECE are evaluated on log binning scheme. }
\label{tab:llm_improvement}
\end{table}

\subsection{Calibration error vs.~RT prediction error}
\label{sec:5.3}

\begin{table}[hb]
\setlength{\tabcolsep}{2pt} 
\centering
\small
\begin{tabular}{llrrcc}
\hline
&        & ECE$\downarrow$   & $\mathrm{ECE}_{\mathrm{log}}\downarrow$& CECE$\downarrow$     & $\mathrm{CECE}_{\mathrm{log}}\downarrow$ \\ \hline
\multirow{2}{*}{Dundee} & $T = 1$  & 1.43  & 1.59 & 4.05E-03 & 4.07E-03   \\
& $T = T^{*}$ & 28.68 & 28.68 &  7.30E-03 & 9.88E-03   \\ \hline
\multirow{2}{*}{NS} & $T = 1$  & 2.48  & 1.91 & 1.83E-02 & 1.53E-02   \\
& $T = T^{*}$ & 35.85 & 35.85 & 3.16E-02 & 3.97E-02   \\ \hline
\multirow{2}{*}{Brown} & $T = 1$  & 1.82  & 1.69 & 1.67E-02 & 1.53E-02   \\
& $T = T^{*}$ & 33.16 & 33.16 & 2.75E-02 & 3.34E-02   \\ \hline
\end{tabular}
\caption{Expected calibration errors (\% ECE and \% CECE) for GPT-2 small on Dundee, Natural Stories (NS) and Brown. Results are all evaluated on the equally-spaced binning scheme and log binning scheme. }
\label{tab:cal_table}
\end{table}

Table~\ref{tab:cal_table} shows ECE and CECE in both equally-spaced and log binning schemes when $T$ equals $1$ and $T^{*}$ on three corpora.
Probability distribution shaped by an optimal $T^{*}$ learnt for fit to human RTs drastically hurts the model calibration regarding these two metrics. 
ECE and CECE with $T^{*}$ are more than 10 times worse than values with $T = 1$. 
This discrepancy can be attributed to the different minima of deviations in LM human RT prediction and expected calibration error. 
The former is minimized towards words where LMs surprisal significantly deviates from human processing difficulty, while the latter is typically minimized with respect to the negative log-likelihood on a hold-out dataset~\citep{guo2017calibration, rahimi2020post}. 

\section{Linguistic analysis}
\label{sec:discussion}

Next we want to gain insight into what words benefit the most from temperature scaling. To this end, we analyze residuals from fitting LME regression models, identifying data points where scaling the temperature parameter notably enhances the fit of human RTs. 
Specifically, we quantify the  improvement in fit by comparing the mean squared error (MSE) before and after adjusting the temperature to its optimal value as follows:
\begin{align}
\Delta_{\mathrm{MSE}}(F) = \mathrm{MSE}_{T = 1}(x_F) - \mathrm{MSE}_{T = T^{*}}(x_F),
\end{align}
where $\mathrm{MSE}_{T = T^{'}}(x_F)$ is the MSE calculated by all the data $x_{F}$ under the linguistic factor $F$. 
The difference $\Delta_{\mathrm{MSE}}(F)$ thus quantifies the impact of scaling relative to the linguistic factor $F$. A higher $\Delta_{\mathrm{MSE}}(F)$ signifies a greater influence of temperature-scaled surprisal of factor $F$. 
To ensure sufficient data in each subset, we only consider subsets including more than 1\% of the data in each corpus. 

\subsection{Influence of low probability words}
\begin{figure}[]
\centering 
\hspace{-5mm}
\hspace{-5mm}
\subfigure{
\includegraphics[width=0.4\textwidth]{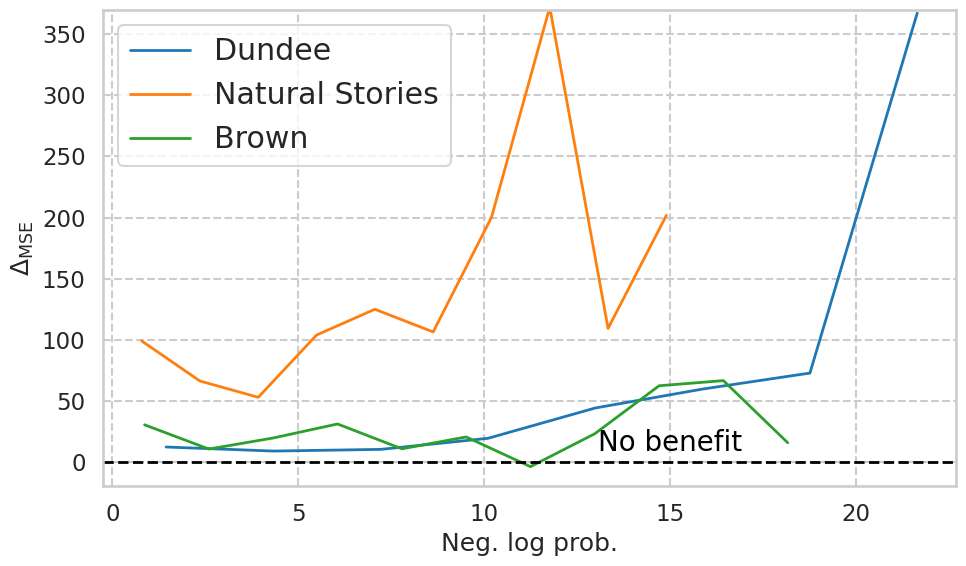}}
\hspace{-5mm}
\hspace{-5mm}
\vspace{-2mm}\\
\caption{Relationship between $\Delta_{\mathrm{MSE}}$ and negative log actual-word probability (surprisal). We take the number of bins to 20. Black dashed lines denote $\Delta_{\mathrm{MSE}} = 0$. Subsets containing less than 1\% of data are ignored for each corpus. }
\label{fig:analysis_probs}
\end{figure}

Given that temperature scaling enhances human likeness by shaping the probability distribution,
it is natural to think about investigating whether there exists an inherent relationship between the distribution of probability 
and $\Delta_{\mathrm{MSE}}$. Specifically, one might ask questions like if samples with low probability gain more from temperature scaling or the other way around.  We find that high surprisal words benefit more from temperature scaling than low surprisal words, across all corpora, see Fig.~\ref{fig:analysis_probs}.  

\subsection{Influence of word types}
We investigate the effects of word-level properties, which include:
\label{sec:word-level_properties}

\textbf{Named entities}. 
Research has substantiated that named entities (NEs) require increased reading time for humans since during the processing of such words~\citep{damasio2004neural, wang2013recognizing}. 
\citet{oh2023does} showed that NEs are among the top two significant factors contributing to the discrepancies of large and small LMs across all corpus-by-LM combinations.
Therefore, we were wondering whether the effect of temperature-scaling might be driven by NE.
To test this, we automatically tagged NEs using a BERT base model~\citep{devlin-etal-2019-bert} fined-tuned for NER
\footnote{Link: https://huggingface.co/dslim/bert-base-NER}. 

\textbf{Part-of-speech tags}. 
Similarly, previous research has argued that the poor fit of large LMs is primarily due to assigning too low surprisal estimates to open-class words like nouns and adjectives~\citep{oh2023does}. 
We POS-tagged the corpora using the NLTK toolkit~\citep{bird2009natural} with the default Penn Treebank Tag set. 
In the following, we mainly focus on the four classes of open-class tags, as well as a subset of the whole closed-class tags (CC). 



\begin{table}[h]
\setlength{\tabcolsep}{2.2pt} 
\centering
\scriptsize
\begin{tabular}{ccr|rr|rrrrr}
\hline
\multicolumn{1}{l}{}                                  &                           &                & \multicolumn{2}{c|}{Named entities} & \multicolumn{5}{c}{POS tags}                                  
\\ \hline
\multicolumn{1}{c|}{}                           & \multicolumn{1}{c|}{GPT2} & Avg.           & NE                & non-NE
& NN          & ADJ         & VERB        & ADV  & CC
\\ \hline
\multicolumn{1}{c|}{\multirow{4}{*}{Dundee}}          & \multicolumn{1}{l|}{s}    & 26.3           & \underline{87.0}         & 23.4           & \underline{33.8}  & \underline{100.5} & -2.0          & 2.6  & 10.4         
\\
\multicolumn{1}{c|}{}                                 & \multicolumn{1}{l|}{m}    & 41.7  & \underline{152.3}        & 36.4           & \underline{57.0}  & \underline{123.3} & 7.8         & 27.6 & 16.4         
\\
\multicolumn{1}{c|}{}                                 & \multicolumn{1}{l|}{l}    & 40.1           & \underline{158.2}        & 34.5           & \underline{56.3}  & \underline{126.5} & 4.8         & 19.2 & 14.0         
\\
\multicolumn{1}{c|}{}                                 & \multicolumn{1}{l|}{xl}   & 41.4  & \underline{168.2}        & 35.4           & \underline{60.0}  & \underline{125.5} & 6.9         & 19.7 & 13.5         
\\ \hline
\multicolumn{1}{c|}{\multirow{4}{*}{NS}} & \multicolumn{1}{l|}{s}    & 105.7          & \underline{186.8}        & 104.6          & \underline{148.7} & \underline{152.5} & 122.0         & 49.0   & 77.1         
\\
\multicolumn{1}{c|}{}                                 & \multicolumn{1}{l|}{m}    & 108.5          & \underline{155.9}        & 107.9          & \underline{145.3} & \underline{152.0} & 130.1       & 60.8 & 80.8         
\\
\multicolumn{1}{c|}{}                                 & \multicolumn{1}{l|}{l}    & 127.7 & \underline{151.6}        & 127.3          & \underline{175.6} &\underline{158.6} & 152.9       & 74.8 & 94.3         
\\
\multicolumn{1}{c|}{}                                 & \multicolumn{1}{l|}{xl}   & 123.3 & \underline{141.8}        & 123.1          & \underline{163.6} & 145.4       & \underline{161.2} & 81.5 & 89.0         
\\ \hline
\multicolumn{1}{c|}{\multirow{4}{*}{Brown}}           & \multicolumn{1}{l|}{s}    & 37.2           & \underline{266.0}        & 28.1           & \underline{54.3}        & -65.2       & \underline{138.1} & 32.1 & 5.9          
\\
\multicolumn{1}{c|}{}                                 & \multicolumn{1}{l|}{m}    & 41.4           & \underline{257.6}        & 32.8           & \underline{71.4}        & -60.6       & \underline{137.5} & 38.6 & 3.5          
\\
\multicolumn{1}{c|}{}                                 & \multicolumn{1}{l|}{l}    & 42.6           & \underline{265.3}        & 51.1           & \underline{69.9}        & -110.3      & \underline{160.8} & 17.2 & 24.7         
\\
\multicolumn{1}{c|}{}                                 & \multicolumn{1}{l|}{xl}   & 54.8  & \underline{282.3}        & 45.8           & \underline{90.5}        & -90.2       & \underline{151.3} & 32.2 & 20.0         
\\ \hline
\end{tabular}
\caption{$\Delta_{\mathrm{MSE}}$ measurement on word-level properties of GPT-2 models on Dundee, Natural Stories (NS) and Brown. Top-3 on each corpus-by-LM are underlined.} 
\label{tab:word-level_properties1}
\end{table}

\textbf{Results}. The result, as shown in Table~\ref{tab:word-level_properties1}, shows primary factors responsible for the benefit of using $s_{T}(w_{t}, T)$ for each corpus-by-LM combination. 
The top three influential subsets for each corpus are underlined. 
Among all datasets and models, \textbf{named entities perform to be the most beneficial word-level attribute.}
In contrast, \textbf{closed-class words profit the least from temperature scaling. } 
Performance trends are consistent across different model variants on the same corpus.

We also measured empirically how often temperature scaling increased vs.~decreased the surprisal estimate of a word. Our results show that for ca.~90\% of words, surprisal estimates are increased through temperature scaling across all word classes. For the subset of named entities, a slightly smaller percentage exhibits increased surprisal estimates. For a full analysis across different corpora and models, see Table \ref{tab:word-level_properties2} in Appendix~\ref{sec:appx_word-level_properties}.

We further investigate the benefit of temperature-scaled surprisal (quantified by $\Delta_{\mathrm{MSE}}$) given the subset of words whose probability decreases (or increases). 
The results are in Table~\ref{tab:word-level_properties3}. 
On Dundee, the main gain arises from the reduction of large probabilities via temperature scaling. 
Conversely, for Natural Stories, the primary benefit comes more strongly from words with originally very low probability, which become more probable. 
For Brown, the effects are evenly split. 
These findings align with our theoretical intuition that \textbf{temperature scaling enhances the fit performance by making probabilities more smooth}, which means not only making high probabilities lower but also making very low probabilities higher and close to $1/K$, since a very low probability also means the model is confident in the incorrectness of certain classes. 

Considering effects on named entities more specifically, we find that on Natural Stories and Brown, the benefit of temperature scaling can mostly be attributed to reducing the probability estimates of highly predictable entities, while on Dundee the beneficial effect mostly arises from increasing probabilities of named entities. We speculate that this could be due to the types of most frequent named entities that occur in the different text sorts, and present a more detailed analysis of this aspect in Appendix ~\ref{sec:appx_word-level_properties}.

\begin{table}[]
\setlength{\tabcolsep}{2.5pt} 
\centering
\small
\begin{tabular}{cc|rr|rr|rr}
\hline
\multicolumn{1}{l}{}              &      &               &                & \multicolumn{4}{c}{Named entities}      \\
\multicolumn{1}{l}{}              &      & \multicolumn{2}{c|}{Avg.}       & \multicolumn{2}{c|}{NE}         & \multicolumn{2}{c}{non-NE}                           \\
      Corpus       & GPT2 & $p_{w_t} {\downarrow}$ & $p_{w_t} {\uparrow}$   & $p_{w_t} {\downarrow}$  & $p_{w_t} {\uparrow} ^{*}$  & $p_{w_t} {\downarrow}$                       & $p_{w_t} {\uparrow}$   \\ \hline
& s    & \textbf{27.4} & 18.2           & 81.3           & \textbf{107.2} & {\color[HTML]{000000} \textbf{25.1}} & 10.1           \\
& m    & \textbf{41.9} & 39.8           & 139.1          & \textbf{205.6} & {\color[HTML]{000000} \textbf{37.8}} & 23.9           \\
& l    & \textbf{41.0} & 31.3           & 156.1          & \textbf{166.6} & {\color[HTML]{000000} \textbf{36.2}} & 18.0           \\
\multirow{-4}{*}{Dundee}          & xl   & \textbf{42.5} & 29.8           & \textbf{170.2} & 158.8          & {\color[HTML]{000000} \textbf{37.0}} & 16.9           \\ \hline
& s    & 94.5          & \textbf{275.6} & \textbf{218.5} & 3.0            & {\color[HTML]{000000} 92.9}          & \textbf{284.9} \\
& m    & 105.7         & \textbf{158.3} & \textbf{179.3} & -34.9          & {\color[HTML]{000000} 104.7}         & \textbf{163.9} \\
& l    & 125.0           & \textbf{166.1} & \textbf{197.5} & -224.8         & {\color[HTML]{000000} 124}           & \textbf{175.4} \\
\multirow{-4}{*}{NS}           & xl   & 121.8         & \textbf{140.7} & \textbf{197.3} & -272.6         & {\color[HTML]{000000} 120.8}         & \textbf{149.5} 
\\ \hline
& s    & \textbf{37.6} & 32.6           & \textbf{329.7} & -170.6         & {\color[HTML]{000000} 26.6}          & \textbf{45.5}  \\
& m    & 39.1          & \textbf{72.3}  & \textbf{276.0}   & 143.6          & {\color[HTML]{000000} 30.5}          & \textbf{66.3}  \\
& l    & \textbf{52.7} & 28.1           & \textbf{325.8} & -205.9         & {\color[HTML]{000000} 42.5}          & \textbf{44.4}  \\
\multirow{-4}{*}{Brown}           & xl   & 50.9          & \textbf{111.5} & \textbf{298.2} & 168.2          & {\color[HTML]{000000} 41.7}          & \textbf{107.1} \\ \hline
\end{tabular}
\caption{Given words whose probability decreases (and increases), the corresponding $\Delta_{\mathrm{MSE}}(p_{w_t} {\downarrow})$ (and $\Delta_{\mathrm{MSE}}(p_{w_t} {\uparrow})$) measurement for GPT-2 models on Dundee, Natural Stories (NS) and Brown. 
A higher $\Delta_{\mathrm{MSE}}$ is displayed in bold in the average across all word types (Avg.), named entities (NE), and non-named entities (non-NE) columns, respectively, for each corpus-by-LM combination. The column with $*$ indicates insufficient (less than 1\%) data. }
\label{tab:word-level_properties3}
\end{table}

\subsection{Influence of multiple-token words}
\label{sec:multiple-token_words}
A fact that is often ignored \citep[but see][]{nair2023words} is that modern LLMs use subword tokenization. This means that long words may consist of several tokens. In this case, the probability of the complete word is calculated by multiplying the probabilities of the subword tokens (and the word's surprisal is correspondingly calculated by adding the surprisals of the subwords). While this may often not matter, whether a word is tokenized into a single subword or several subwords can make a remarkable difference when applying temperature scaling: imagine a long / difficult word which has a low probability (and correspondingly a high surprisal). If this word were to be represented as a single subword token, temperature scaling might have the effect that the probability of this word gets \textit{increased} during temperature scaling, and its surprisal estimate is hence decreased at $T>1$.


If, on the other hand, the same word were to be composed of two subword tokens, one or both of the subword tokens can be expected to have a higher probability (than a hypothetical single subword token), and it is possible that during temperature scaling, the probabilities of the subword tokens would each be \textit{decreased} at $T>1$, such that the sum of the surprisals of the subword tokens would be much higher, compared to the word's surprisal estimate at $T=1$. 

To summarize, whether the surprisal of a certain word would increase or decrease after temperature scaling could depend on whether that word happens to be included in the subword token vocabulary or not.\footnote{Distributions of surprisal for single vs.~multiple token words before and after temperature scaling are provided in Fig.~\ref{fig:prob_dis} in Appendix~\ref{sec:appx_attached_figs_tables}.}
In order to quantify to what extent subword tokenization affects surprisal estimates, we conducted several analyses. 

\begin{figure}[]
\centering 
\hspace{-15mm}
\subfigure{
\includegraphics[width=0.49\textwidth]{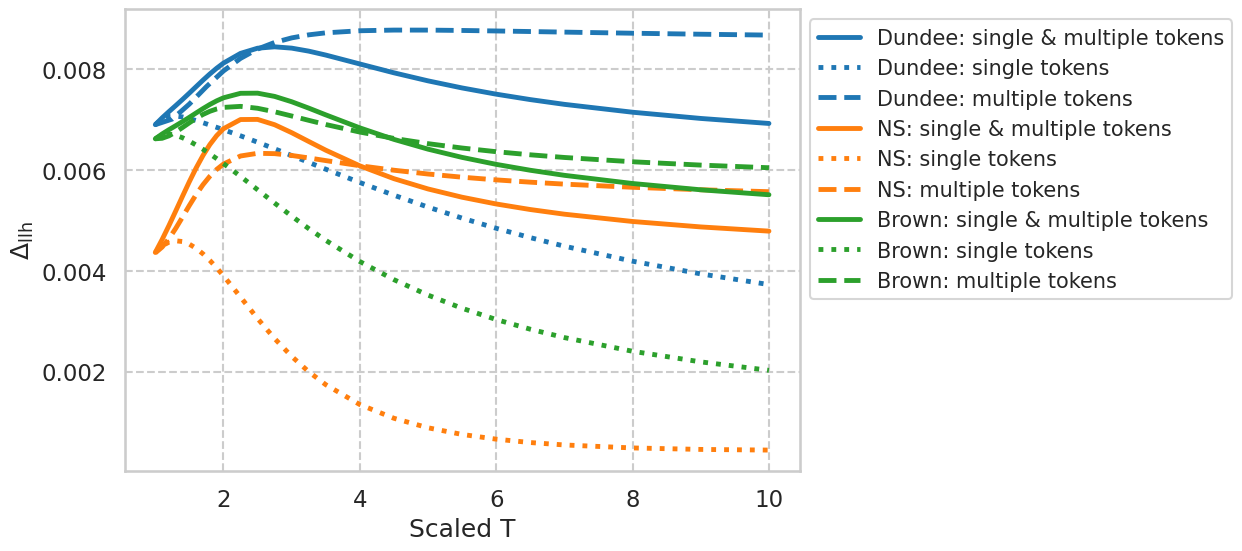}}
\hspace{-15mm}
\vspace{-5mm}
\caption{Relationship between $\Delta_{\mathrm{llh}}$ of GPT-2 s on three corpora and corresponding temperature T. 
}
\vspace{-5mm}
\label{fig:multitoken_deltallh}
\end{figure}

Fig.~\ref{fig:multitoken_deltallh} shows $\Delta_{\mathrm{llh}}$ under various conditions:  scaling all words (consistent with experiments in Section~\ref{sec:results1}) vs.~taking into the analysis only the subset of single-token words and multiple-token words.
The comparison between the full, dotted, and dashed lines highlights that \textbf{the benefit of temperature-scaled surprisal comes primarily from the scaling of multiple-token words}. 

Next, it is interesting to consider for what percentage of multiple-token words temperature-scaling \textit{increases} the surprisal. We find that the surprisal of more than 90\% of multiple-token words increases, and the \textbf{ratio is higher than across single-token words} by ca.~6\% on Dundee and Brown, see
Table~\ref{tab:influence_of_multipletoken_words} in Appendix~\ref{sec:appx_attached_figs_tables} for more details.



\section{Discussion}
Our experiments demonstrate that choosing a temperature around 2.5 improves the fit to human reading times. Furthermore, we find that this effect is chiefly driven by an improved fit for words which consist of several subword tokens.\footnote{Appendix~\ref{appx:ne-vs-multitok} shows that subword tokenization has larger explanatory power than word class.} Named entities and other open class words tend to have a larger tendency to contain several subword tokens, which can explain why temperature scaling is particularly effective for these words. 


So what does all of this mean for surprisal estimates from LLMs and reading time prediction? Firstly, following the argumentation of \citet{oh2023does}, it is possible that indeed the effect is driven by humans failing to accurately estimate the probability of rare words (rare words being the ones that are split up into several subwords), because they do not reach sufficient language experience or because human language models do not track these probabilities well. In this case, temperature-scaling rare words to which the LLM assigns a too high probability (and hence a low surprisal) would be a good strategy to counteract the discrepancy between humans and LLMs.
From LLMs' perspective, recalling the observation from Section~\ref{sec:5.3} that larger LLMs that yield poorer fits to RTs are actually better calibrated, hence the massive training dataset might be at the cause of driving these models away from the human-like predictive processing, aligning with~\citet{Oh_Schuler_23}.

Secondly, it is likely that the beneficial effect of temperature scaling is an artifact of subword tokenization, and that this effect would diminish if all words were composed of only a single subword token (cf.~our explanation in Section \ref{sec:multiple-token_words}). That is, temperature scaling would not be beneficial because of the reasons that motivated this research originally, but only because it is a way of assigning higher surprisal to words consisting of several subword tokens.
In order to test this hypothesis, one would have to re-train a GPT-2 model using a vocabulary that at least includes all words that are contained in the reading time corpora, and then re-running the analysis to check whether a beneficial effect of temperature scaling can still be found.

Finally, it is also possible that the splitting of a word into subwords coincides with the reader fixating a word several times, and that these added fixations lead to an overestimate in RTs compared to the actual surprisal experienced by a human reader. Future work could investigate this hypothesis by analysing RTs on subwords instead of aggregated words (with the caveat that subword tokens may not be cognitively plausible units). 

\section{Conclusion}
\label{sec:conclusion}
This paper studies the prediction of human RTs from the perspective of probability distribution. We make the following contributions: (1) We demonstrate that the prediction of RTs can be significantly improved via temperature scaling of LLM probability estimates. (2) We demonstrate that the primary benefit of temperature-scaled surprisal is driven by words composed of several subword tokens. These words also tend to be rarer / long open-class words. 
Future work should investigate the interaction of subword tokenization and temperature scaling, as well as the issue of tokenization in the analysis of eye-tracking data.

\section*{Limitations} 
In this work, the identification of the optimal T for temperature-scaled surprisal is manually tuned. 
Future research could develop an automated method to determine this optimal value, e.g., from specific characteristics of LLMs or corpora. 
Additionally, a question may be asked whether the possible non-linear relationship between surprisal and reading times~\citep{shain2022large, hoover2023plausibility} could influence the temperature-scaled surprisal’s superiority over original surprisal. 
Investigating the effectiveness of temperature-scaled surprisal using generalized additive models, a branch of models that assume less about the linearity than linear mixed effect models employed here, would be an extension. 
Finally, exploring effects of temperature-scaled surprisal on different measures of fixation duration could be considered in future work. 

\section*{Ethical Considerations}
The datasets and packages we used are all publicly available and have no privacy issues.

\section*{Acknowledgements}
The authors thank Xudong Hong and Dongqi Pu for useful discussions and comments. 

\bibliography{anthology}

\begin{thebibliography}{53}
\expandafter\ifx\csname natexlab\endcsname\relax\def\natexlab#1{#1}\fi

\bibitem[{Aaronson and Scarborough(1976)}]{aaronson1976performance}
Doris Aaronson and Hollis~S Scarborough. 1976.
\newblock \href {https://psycnet.apa.org/record/1976-08801-001} {Performance theories for sentence coding: Some quantitative evidence.}
\newblock \emph{Journal of Experimental Psychology: Human perception and performance}, 2(1):56.

\bibitem[{Angele et~al.(2015)Angele, Schotter, Slattery, Tenenbaum, Bicknell, and Rayner}]{angele2015successor}
Bernhard Angele, Elizabeth~R Schotter, Timothy~J Slattery, Tara~L Tenenbaum, Klinton Bicknell, and Keith Rayner. 2015.
\newblock \href {https://www.sciencedirect.com/science/article/pii/S0749596X1400151X} {Do successor effects in reading reflect lexical parafoveal processing? evidence from corpus-based and experimental eye movement data}.
\newblock \emph{Journal of Memory and Language}, 79:76--96.

\bibitem[{Aurnhammer and Frank(2019)}]{aurnhammer2019evaluating}
Christoph Aurnhammer and Stefan~L Frank. 2019.
\newblock \href {https://www.sciencedirect.com/science/article/pii/S0028393219302404} {Evaluating information-theoretic measures of word prediction in naturalistic sentence reading}.
\newblock \emph{Neuropsychologia}, 134:107198.

\bibitem[{Bates et~al.(2015)Bates, M{\"a}chler, Bolker, and Walker}]{bates2015fitting}
Douglas Bates, Martin M{\"a}chler, Ben Bolker, and Steve Walker. 2015.
\newblock Fitting linear mixed-effects models using lme4.
\newblock \emph{Journal of Statistical Software}, 67:1--48.

\bibitem[{Bird et~al.(2009)Bird, Klein, and Loper}]{bird2009natural}
Steven Bird, Ewan Klein, and Edward Loper. 2009.
\newblock \href {https://thuvienso.dau.edu.vn:88/bitstream/DHKTDN/6460/1/Natural%20Language%20Processing%20with%20Python.4149.pdf} {\emph{Natural language processing with Python: analyzing text with the natural language toolkit}}.
\newblock " O'Reilly Media, Inc.".

\bibitem[{Caccia et~al.(2018)Caccia, Caccia, Fedus, Larochelle, Pineau, and Charlin}]{caccia2018language}
Massimo Caccia, Lucas Caccia, William Fedus, Hugo Larochelle, Joelle Pineau, and Laurent Charlin. 2018.
\newblock \href {https://arxiv.org/pdf/1811.02549.pdf} {Language gans falling short}.
\newblock \emph{arXiv preprint arXiv:1811.02549}.

\bibitem[{Damasio et~al.(2004)Damasio, Tranel, Grabowski, Adolphs, and Damasio}]{damasio2004neural}
Hanna Damasio, Daniel Tranel, Thomas Grabowski, Ralph Adolphs, and Antonio Damasio. 2004.
\newblock \href {https://www.sciencedirect.com/science/article/pii/S0010027703002312?casa_token=33sljq6jIZEAAAAA:Pf_G0jh3BpzUIPquX92jFSd-K-yqEZVZUcdQ3AV_PFXBeuDw82TefiDVc5ZRNIhcB1Zsmd50RA} {Neural systems behind word and concept retrieval}.
\newblock \emph{Cognition}, 92(1-2):179--229.

\bibitem[{Demberg and Keller(2008)}]{demberg2008data}
Vera Demberg and Frank Keller. 2008.
\newblock \href {https://www.sciencedirect.com/science/article/pii/S0010027708001741?casa_token=6CfFPlamqUMAAAAA:0NOXAwKkL-oHz18OECcqqIPZwI8smBD4Qobv3_sF96w3FUR46efOKceJtSDC2CbFSJwpNrq3NA} {Data from eye-tracking corpora as evidence for theories of syntactic processing complexity}.
\newblock \emph{Cognition}, 109(2):193--210.

\bibitem[{Devlin et~al.(2019)Devlin, Chang, Lee, and Toutanova}]{devlin-etal-2019-bert}
Jacob Devlin, Ming-Wei Chang, Kenton Lee, and Kristina Toutanova. 2019.
\newblock \href {https://doi.org/10.18653/v1/N19-1423} {{BERT}: Pre-training of deep bidirectional transformers for language understanding}.
\newblock In \emph{Proceedings of the 2019 Conference of the North {A}merican Chapter of the Association for Computational Linguistics: Human Language Technologies, Volume 1 (Long and Short Papers)}, pages 4171--4186, Minneapolis, Minnesota. Association for Computational Linguistics.

\bibitem[{Fan et~al.(2018)Fan, Lewis, and Dauphin}]{fan2018hierarchical}
Angela Fan, Mike Lewis, and Yann Dauphin. 2018.
\newblock \href {https://doi.org/10.18653/v1/P18-1082} {Hierarchical neural story generation}.
\newblock In \emph{Proceedings of the 56th Annual Meeting of the Association for Computational Linguistics (Volume 1: Long Papers)}, pages 889--898, Melbourne, Australia. Association for Computational Linguistics.

\bibitem[{Ficler and Goldberg(2017)}]{ficler2017controlling}
Jessica Ficler and Yoav Goldberg. 2017.
\newblock \href {https://doi.org/10.18653/v1/W17-4912} {Controlling linguistic style aspects in neural language generation}.
\newblock In \emph{Proceedings of the Workshop on Stylistic Variation}, pages 94--104, Copenhagen, Denmark. Association for Computational Linguistics.

\bibitem[{Forster et~al.(2009)Forster, Guerrera, and Elliot}]{forster2009maze}
Kenneth~I Forster, Christine Guerrera, and Lisa Elliot. 2009.
\newblock \href {https://link.springer.com/article/10.3758/brm.41.1.163} {The maze task: Measuring forced incremental sentence processing time}.
\newblock \emph{Behavior research methods}, 41:163--171.

\bibitem[{Fossum and Levy(2012)}]{fossum2012sequential}
Victoria Fossum and Roger Levy. 2012.
\newblock \href {https://aclanthology.org/W12-1706} {Sequential vs. hierarchical syntactic models of human incremental sentence processing}.
\newblock In \emph{Proceedings of the 3rd Workshop on Cognitive Modeling and Computational Linguistics ({CMCL} 2012)}, pages 61--69, Montr{\'e}al, Canada. Association for Computational Linguistics.

\bibitem[{Frank and Bod(2011)}]{frank2011insensitivity}
Stefan~L Frank and Rens Bod. 2011.
\newblock \href {https://journals.sagepub.com/doi/full/10.1177/0956797611409589?casa_token=-TqMYhYCHwgAAAAA%3AmtY-SRlgH-s67iMCuS_5L4EWau2LY221Sj2HFqrh-mFm0MfHH79EgtkrDqGSVpl-sPA4RNLEAWPq} {Insensitivity of the human sentence-processing system to hierarchical structure}.
\newblock \emph{Psychological science}, 22(6):829--834.

\bibitem[{Futrell et~al.(2018)Futrell, Gibson, Tily, Blank, Vishnevetsky, Piantadosi, and Fedorenko}]{futrell2018natural}
Richard Futrell, Edward Gibson, Harry~J Tily, Idan Blank, Anastasia Vishnevetsky, Steven Piantadosi, and Evelina Fedorenko. 2018.
\newblock \href {https://aclanthology.org/L18-1012.pdf} {The natural stories corpus}.
\newblock In \emph{Proceedings of the Eleventh International Conference on Language Resources and Evaluation (LREC 2018)}.

\bibitem[{Goodkind and Bicknell(2018)}]{goodkind2018predictive}
Adam Goodkind and Klinton Bicknell. 2018.
\newblock \href {https://doi.org/10.18653/v1/W18-0102} {Predictive power of word surprisal for reading times is a linear function of language model quality}.
\newblock In \emph{Proceedings of the 8th Workshop on Cognitive Modeling and Computational Linguistics ({CMCL} 2018)}, pages 10--18, Salt Lake City, Utah. Association for Computational Linguistics.

\bibitem[{Guo et~al.(2017)Guo, Pleiss, Sun, and Weinberger}]{guo2017calibration}
Chuan Guo, Geoff Pleiss, Yu~Sun, and Kilian~Q Weinberger. 2017.
\newblock \href {https://proceedings.mlr.press/v70/guo17a/guo17a.pdf} {On calibration of modern neural networks}.
\newblock In \emph{International conference on machine learning}, pages 1321--1330. PMLR.

\bibitem[{Hale(2001)}]{hale2001probabilistic}
John Hale. 2001.
\newblock \href {https://aclanthology.org/N01-1021.pdf} {A probabilistic earley parser as a psycholinguistic model}.
\newblock In \emph{Second meeting of the north american chapter of the association for computational linguistics}.

\bibitem[{Hale(2003)}]{hale2003information}
John Hale. 2003.
\newblock \href {https://link.springer.com/article/10.1023/A:1022492123056} {The information conveyed by words in sentences}.
\newblock \emph{Journal of psycholinguistic research}, 32:101--123.

\bibitem[{Hale(2006)}]{hale2006uncertainty}
John Hale. 2006.
\newblock \href {https://onlinelibrary.wiley.com/doi/abs/10.1207/s15516709cog0000_64} {Uncertainty about the rest of the sentence}.
\newblock \emph{Cognitive science}, 30(4):643--672.

\bibitem[{Hendrycks et~al.(2019)Hendrycks, Lee, and Mazeika}]{hendrycks2019using}
Dan Hendrycks, Kimin Lee, and Mantas Mazeika. 2019.
\newblock \href {https://proceedings.mlr.press/v97/hendrycks19a/hendrycks19a.pdf} {Using pre-training can improve model robustness and uncertainty}.
\newblock In \emph{International conference on machine learning}, pages 2712--2721. PMLR.

\bibitem[{Hinton et~al.(2015)Hinton, Vinyals, and Dean}]{hinton2015distilling}
Geoffrey Hinton, Oriol Vinyals, and Jeff Dean. 2015.
\newblock \href {https://arxiv.org/pdf/1503.02531.pdf} {Distilling the knowledge in a neural network}.
\newblock \emph{arXiv preprint arXiv:1503.02531}.

\bibitem[{Hoover et~al.(2023)Hoover, Sonderegger, Piantadosi, and O’Donnell}]{hoover2023plausibility}
Jacob~Louis Hoover, Morgan Sonderegger, Steven~T Piantadosi, and Timothy~J O’Donnell. 2023.
\newblock \href {https://direct.mit.edu/opmi/article/doi/10.1162/opmi_a_00086/116522} {The plausibility of sampling as an algorithmic theory of sentence processing}.
\newblock \emph{Open Mind}, 7:350--391.

\bibitem[{Kennedy et~al.(2003)Kennedy, Hill, and Pynte}]{kennedy2003dundee}
Alan Kennedy, Robin Hill, and Jo{\"e}l Pynte. 2003.
\newblock The dundee corpus.
\newblock In \emph{Proceedings of the 12th European Conference on Eye Movement}.

\bibitem[{Kull et~al.(2019)Kull, Perello~Nieto, K{\"a}ngsepp, Silva~Filho, Song, and Flach}]{kull2019beyond}
Meelis Kull, Miquel Perello~Nieto, Markus K{\"a}ngsepp, Telmo Silva~Filho, Hao Song, and Peter Flach. 2019.
\newblock \href {https://proceedings.neurips.cc/paper/2019/file/8ca01ea920679a0fe3728441494041b9-Paper.pdf} {Beyond temperature scaling: Obtaining well-calibrated multi-class probabilities with dirichlet calibration}.
\newblock \emph{Advances in neural information processing systems}, 32.

\bibitem[{Kumar et~al.(2019)Kumar, Liang, and Ma}]{kumar2019verified}
Ananya Kumar, Percy~S Liang, and Tengyu Ma. 2019.
\newblock \href {https://proceedings.neurips.cc/paper/2019/file/f8c0c968632845cd133308b1a494967f-Paper.pdf} {Verified uncertainty calibration}.
\newblock \emph{Advances in Neural Information Processing Systems}, 32.

\bibitem[{Kumar(2022)}]{kumar2022answer}
Sawan Kumar. 2022.
\newblock \href {https://doi.org/10.18653/v1/2022.acl-long.49} {Answer-level calibration for free-form multiple choice question answering}.
\newblock In \emph{Proceedings of the 60th Annual Meeting of the Association for Computational Linguistics (Volume 1: Long Papers)}, pages 665--679, Dublin, Ireland. Association for Computational Linguistics.

\bibitem[{Kuribayashi et~al.(2022)Kuribayashi, Oseki, Brassard, and Inui}]{kuribayashi2022context}
Tatsuki Kuribayashi, Yohei Oseki, Ana Brassard, and Kentaro Inui. 2022.
\newblock \href {https://doi.org/10.18653/v1/2022.emnlp-main.712} {Context limitations make neural language models more human-like}.
\newblock In \emph{Proceedings of the 2022 Conference on Empirical Methods in Natural Language Processing}, pages 10421--10436, Abu Dhabi, United Arab Emirates. Association for Computational Linguistics.

\bibitem[{Levy(2008)}]{levy2008expectation}
Roger Levy. 2008.
\newblock \href {https://www.sciencedirect.com/science/article/pii/S0010027707001436?casa_token=YFQzL2RFTqkAAAAA:VQyrXnoLZqHqIouZA0Ui7DQgzfC9DAPECT6McgAk8LUmyMgHXpHmHZZIdu5grZmdG0HDMYtP1g} {Expectation-based syntactic comprehension}.
\newblock \emph{Cognition}, 106(3):1126--1177.

\bibitem[{Linzen and Jaeger(2014)}]{linzen2014investigating}
Tal Linzen and T~Florian Jaeger. 2014.
\newblock \href {https://aclanthology.org/W14-2002.pdf} {Investigating the role of entropy in sentence processing}.
\newblock In \emph{Proceedings of the fifth workshop on cognitive modeling and computational linguistics}, pages 10--18.

\bibitem[{Merkx and Frank(2021)}]{merkx-frank-2021-human}
Danny Merkx and Stefan~L. Frank. 2021.
\newblock \href {https://doi.org/10.18653/v1/2021.cmcl-1.2} {Human sentence processing: Recurrence or attention?}
\newblock In \emph{Proceedings of the Workshop on Cognitive Modeling and Computational Linguistics}, pages 12--22, Online. Association for Computational Linguistics.

\bibitem[{Mitchell and Green(1978)}]{mitchell1978effects}
Don~C Mitchell and David~W Green. 1978.
\newblock \href {https://www.tandfonline.com/doi/abs/10.1080/14640747808400689?casa_token=OL71aT-qSEIAAAAA:hfVjpi1R6FEKL77GAR_Tz-tbM1utlrnCGEDXN-SzBzWMvAQ7HByi_6DVQlMfYw-ssAfNdvkVNthbdQ} {The effects of context and content on immediate processing in reading}.
\newblock \emph{The quarterly journal of experimental psychology}, 30(4):609--636.

\bibitem[{Nair and Resnik(2023)}]{nair2023words}
Sathvik Nair and Philip Resnik. 2023.
\newblock \href {https://aclanthology.org/2023.findings-emnlp.752} {Words, subwords, and morphemes: What really matters in the surprisal-reading time relationship?}
\newblock In \emph{Findings of the Association for Computational Linguistics: EMNLP 2023}, pages 11251--11260, Singapore. Association for Computational Linguistics.

\bibitem[{Oh et~al.(2022)Oh, Clark, and Schuler}]{oh2022comparison}
Byung-Doh Oh, Christian Clark, and William Schuler. 2022.
\newblock \href {https://www.frontiersin.org/articles/10.3389/frai.2022.777963/full} {Comparison of structural parsers and neural language models as surprisal estimators}.
\newblock \emph{Frontiers in Artificial Intelligence}, 5:777963.

\bibitem[{Oh and Schuler(2023{\natexlab{a}})}]{Oh_Schuler_23}
Byung{-}Doh Oh and William Schuler. 2023{\natexlab{a}}.
\newblock \href {https://aclanthology.org/2023.findings-emnlp.128} {Transformer-based language model surprisal predicts human reading times best with about two billion training tokens}.
\newblock In \emph{Findings of the Association for Computational Linguistics: {EMNLP} 2023, Singapore, December 6-10, 2023}, pages 1915--1921. Association for Computational Linguistics.

\bibitem[{Oh and Schuler(2023{\natexlab{b}})}]{oh2023does}
Byung-Doh Oh and William Schuler. 2023{\natexlab{b}}.
\newblock \href {https://doi.org/10.1162/tacl_a_00548} {Why does surprisal from larger transformer-based language models provide a poorer fit to human reading times?}
\newblock \emph{Transactions of the Association for Computational Linguistics}, 11:336--350.

\bibitem[{Pimentel et~al.(2023)Pimentel, Meister, Wilcox, Levy, and Cotterell}]{pimentel-etal-2023-effect}
Tiago Pimentel, Clara Meister, Ethan~G. Wilcox, Roger Levy, and Ryan Cotterell. 2023.
\newblock \href {https://arxiv.org/abs/2211.14301} {On the effect of anticipation on reading times}.
\newblock \emph{Transactions of the Association for Computational Linguistics}.

\bibitem[{Radford et~al.(2019)Radford, Wu, Child, Luan, Amodei, Sutskever et~al.}]{radford2019language}
Alec Radford, Jeffrey Wu, Rewon Child, David Luan, Dario Amodei, Ilya Sutskever, et~al. 2019.
\newblock \href {https://insightcivic.s3.us-east-1.amazonaws.com/language-models.pdf} {Language models are unsupervised multitask learners}.
\newblock \emph{OpenAI blog}, 1(8):9.

\bibitem[{Rahimi et~al.(2020{\natexlab{a}})Rahimi, Gupta, Ajanthan, Mensink, Sminchisescu, and Hartley}]{rahimi2020post}
Amir Rahimi, Kartik Gupta, Thalaiyasingam Ajanthan, Thomas Mensink, Cristian Sminchisescu, and Richard Hartley. 2020{\natexlab{a}}.
\newblock \href {https://www.researchgate.net/profile/Thalaiyasingam-Ajanthan/publication/342408476_Post-hoc_Calibration_of_Neural_Networks/links/5ef95f0fa6fdcc4ca43a1b2e/Post-hoc-Calibration-of-Neural-Networks.pdf} {Post-hoc calibration of neural networks}.
\newblock \emph{arXiv preprint arXiv:2006.12807}, 2.

\bibitem[{Rahimi et~al.(2020{\natexlab{b}})Rahimi, Shaban, Cheng, Hartley, and Boots}]{rahimi2020intra}
Amir Rahimi, Amirreza Shaban, Ching-An Cheng, Richard Hartley, and Byron Boots. 2020{\natexlab{b}}.
\newblock \href {https://proceedings.neurips.cc/paper/2020/file/9bc99c590be3511b8d53741684ef574c-Paper.pdf} {Intra order-preserving functions for calibration of multi-class neural networks}.
\newblock \emph{Advances in Neural Information Processing Systems}, 33:13456--13467.

\bibitem[{Rayner(1998)}]{rayner1998eye}
Keith Rayner. 1998.
\newblock \href {https://psycnet.apa.org/record/1998-11174-004} {Eye movements in reading and information processing: 20 years of research.}
\newblock \emph{Psychological bulletin}, 124(3):372.

\bibitem[{Reeb and Wolf(2015)}]{reeb2015tight}
David Reeb and Michael~M Wolf. 2015.
\newblock \href {https://ieeexplore.ieee.org/abstract/document/7001656?casa_token=OcXEuthJRWkAAAAA:qWEDYzmkI4mma7ETdqNOsu0fG9dDrOYypmNebZMvcPRZywigOGkNkYaDD1aaT9yH920tgJTEplv_} {Tight bound on relative entropy by entropy difference}.
\newblock \emph{IEEE Transactions on Information Theory}, 61(3):1458--1473.

\bibitem[{R{\'e}nyi(1961)}]{renyi1961measures}
Alfr{\'e}d R{\'e}nyi. 1961.
\newblock \href {https://static.renyi.hu/renyi_cikkek/1961_on_measures_of_entropy_and_information.pdf} {On measures of entropy and information}.
\newblock In \emph{Proceedings of the Fourth Berkeley Symposium on Mathematical Statistics and Probability, Volume 1: Contributions to the Theory of Statistics}, volume~4, pages 547--562. University of California Press.

\bibitem[{Shain et~al.(2022)Shain, Meister, Pimentel, Cotterell, and Levy}]{shain2022large}
Cory Shain, Clara Meister, Tiago Pimentel, Ryan Cotterell, and Roger Levy. 2022.
\newblock \href {https://psyarxiv.com/4hyna/} {Large-scale evidence for logarithmic effects of word predictability on reading time}.

\bibitem[{Shannon(1948)}]{shannon1948mathematical}
Claude~Elwood Shannon. 1948.
\newblock \href {https://ieeexplore.ieee.org/abstract/document/6773024?casa_token=wOUk3OSa4pAAAAAA:aJjxiYCpWL_jQPt52e5hktJp3bCmZ9e00MKiA3-KHjq7GyQuTrx8s1EKQfGt4xdtAVFjQFIUZ87S} {A mathematical theory of communication}.
\newblock \emph{The Bell system technical journal}, 27(3):379--423.

\bibitem[{Si et~al.(2022)Si, Zhao, Min, and Boyd-Graber}]{si2022re}
Chenglei Si, Chen Zhao, Sewon Min, and Jordan Boyd-Graber. 2022.
\newblock \href {https://doi.org/10.18653/v1/2022.findings-emnlp.204} {Re-examining calibration: The case of question answering}.
\newblock In \emph{Findings of the Association for Computational Linguistics: EMNLP 2022}, pages 2814--2829, Abu Dhabi, United Arab Emirates. Association for Computational Linguistics.

\bibitem[{Smith and Levy(2013)}]{smith2013effect}
Nathaniel~J Smith and Roger Levy. 2013.
\newblock \href {https://www.sciencedirect.com/science/article/pii/S0010027713000413} {The effect of word predictability on reading time is logarithmic}.
\newblock \emph{Cognition}, 128(3):302--319.

\bibitem[{Srivastava et~al.(2023)Srivastava, Rastogi, Rao, Shoeb, Abid, Fisch, Brown, Santoro, Gupta, Garriga-Alonso et~al.}]{srivastava2023beyond}
Aarohi Srivastava, Abhinav Rastogi, Abhishek Rao, Abu Awal~Md Shoeb, Abubakar Abid, Adam Fisch, Adam~R Brown, Adam Santoro, Aditya Gupta, Adri{\`a} Garriga-Alonso, et~al. 2023.
\newblock \href {https://arxiv.org/abs/2206.04615} {Beyond the imitation game: Quantifying and extrapolating the capabilities of language models}.
\newblock \emph{Transactions on Machine Learning Research}.

\bibitem[{van Schijndel and Linzen(2019)}]{van2018can}
Marten van Schijndel and Tal Linzen. 2019.
\newblock \href {https://doi.org/10.7275/qtbb-9d05} {Can entropy explain successor surprisal effects in reading?}
\newblock In \emph{Proceedings of the Society for Computation in Linguistics ({SC}i{L}) 2019}, pages 1--7.

\bibitem[{Wang et~al.(2013)Wang, Zhu, Bastiaansen, Hagoort, and Yang}]{wang2013recognizing}
Lin Wang, Zude Zhu, Marcel Bastiaansen, Peter Hagoort, and Yufang Yang. 2013.
\newblock \href {https://www.sciencedirect.com/science/article/pii/S0093934X13000229?casa_token=gMuI4AO-9ewAAAAA:gf74S5ZHqD83XJy-GxZ_z0LpRZtOFPhI9yEAkoUDatoM6VwshPWB_Wrw5FzYUH27LW607Coszw} {Recognizing the emotional valence of names: An erp study}.
\newblock \emph{Brain and Language}, 125(1):118--127.

\bibitem[{Wilcox et~al.(2020)Wilcox, Gauthier, Hu, Qian, and Levy}]{wilcox2020predictive}
Ethan~Gotlieb Wilcox, Jon Gauthier, Jennifer Hu, Peng Qian, and Roger~P. Levy. 2020.
\newblock \href {https://arxiv.org/abs/2006.01912} {On the predictive power of neural language models for human real-time comprehension behavior}.
\newblock In \emph{Proceedings of the 42nd Annual Meeting of the Cognitive Science Society}, page 1707–1713.

\bibitem[{Wilcox et~al.(2023)Wilcox, Pimentel, Meister, Cotterell, and Levy}]{wilcox2023testing}
Ethan~Gotlieb Wilcox, Tiago Pimentel, Clara Meister, Ryan Cotterell, and Roger~P Levy. 2023.
\newblock \href {https://arxiv.org/abs/2307.03667} {Testing the predictions of surprisal theory in 11 languages}.
\newblock \emph{Transactions of the Association for Computational Linguistics}.

\bibitem[{Zhu et~al.(2023)Zhu, Xu, Wang, Zhang, and Mao}]{zhu-etal-2023-calibration}
Chiwei Zhu, Benfeng Xu, Quan Wang, Yongdong Zhang, and Zhendong Mao. 2023.
\newblock \href {https://doi.org/10.18653/v1/2023.findings-emnlp.654} {On the calibration of large language models and alignment}.
\newblock In \emph{Findings of the Association for Computational Linguistics: EMNLP 2023}, pages 9778--9795, Singapore. Association for Computational Linguistics.

\end{thebibliography}

\newpage
\appendix

\section{Connection to Contextual Rényi Entropy}
\label{sec:appx_theorems}
While a lot of work has investigated the effect of next word entropy on reading times \citep{hale2003information, hale2006uncertainty, linzen2014investigating, angele2015successor, van2018can, aurnhammer2019evaluating, pimentel-etal-2023-effect}, we will here focus on contextual R\'enyi entropy (the entropy of the probability distribution at the current time stamp, which is parameterized by $\alpha$), as proposed in \citet{pimentel-etal-2023-effect} to represent human anticipatory reading process. \citet{pimentel-etal-2023-effect} find that Rényi entropy with an optimal $\alpha^{*}$ in the range of $(0, 1)$ (around $1/2$) obtains the best performance in reading time prediction (compared to Shannon Entropy ($\alpha=1$) or compared to unscaled surprisal estimates).

Mathematically, Contextual Rényi entropy~\citep{renyi1961measures} is defined as: 
\begin{align}
\begin{split}
\mathrm{H}_{\alpha}(w_{t}& \mid \boldsymbol{w}_{<t})\\
& =
\lim_{\beta \rightarrow \alpha} \frac{1}{1- \beta}\mathrm{log}_{2} \sum_{w \in \overline{\mathcal{W}}} (p(w|\boldsymbol{w}_{<t}))^{\beta}.\end{split}
\end{align}
For given $\alpha^{'} \in (0, \infty)$, we simply denote $\mathrm{H}_{\alpha}(w_{t} \mid \boldsymbol{w}_{<t})|_{\alpha = \alpha^{'}}$ as $\mathrm{H}_{\alpha}|_{\alpha = \alpha^{'}}$. 

\begin{theorem}
\label{theorem:1}
\emph{(Monotonicity of $s_{T}(w_{t}, T)$ and $\mathrm{H}_{\alpha}(w_{t} \mid \boldsymbol{w}_{<t})$).}
Given any probability distribution $\boldsymbol{p}$ with actual-word probability $p_{w_{t}} > 1/K$, where $K$ is the number of classes, 
temperature-scaled surprisal $s_{T}(w_{t}, T)$ is strictly monotonically increasing in  $\Delta_{T} \in [1, \infty]$, 
Rényi entropy $\mathrm{H}_{\alpha}(w_{t} \mid \boldsymbol{w}_{<t})$ is strictly monotonically decreasing in  $\Delta_{\alpha} \in [0, 1]$, 
especially, 
\begin{align}
s_{T}|_{T=1} < s_{T}|_{T = T^{*}} <  \lim_{T \to \infty} s_{T}(w_{t}, T) \label{eq:theorem1.1}\\
\mathrm{H}_{\alpha}|_{\alpha = 1} <  \mathrm{H}_{\alpha}|_{\alpha = 1/2} < \mathrm{H}_{\alpha}|_{\alpha = 0}, \label{eq:theorem1.2}
\end{align}
where $T^{*}$ is the optimal $T$ of fit to RTs in the range of $\Delta_{T}$. 
\end{theorem}
\textit{Proof.} 
Eq.~\eqref{eq:theorem1.1} can be easily verified by considering the monotonicity of temperature-scaled softmax output $\sigma_{SM}(\hat{\boldsymbol{z}}_{w_{t}}/T)$.
The second part of Eq.~\eqref{eq:theorem1.2} can be rewritten as: 
\begin{align}
\mathrm{H}_{\alpha}|_{\alpha = 1/2} &= 2 \log_{2}\sum_{w \in \overline{\mathcal{W}}} \sqrt{p(w|\boldsymbol{w}_{<t})} \label{eq:theorem1_sub1}\\
&< 2 \log_{2}\sqrt{K \sum_{w \in \overline{\mathcal{W}}}p(w|\boldsymbol{w}_{<t})} \label{eq:theorem1_sub2}\\
&=  -\log_{2}(1/K) = \mathrm{H}_{\alpha}|_{\alpha = 0},\label{eq:theorem1_sub3}
\end{align}
where for the step from Eq.~\eqref{eq:theorem1_sub1} to Eq.~\eqref{eq:theorem1_sub2} we use AM-QM inequality and $K$ is the number of classes in tokenizer.  
The first part of Eq.~\eqref{eq:theorem1.2} can be rewritten as: 
\begin{align}
\mathrm{H}_{\alpha}|_{\alpha = 1/2} &= 2 \log_{2}\sum_{w \in \overline{\mathcal{W}}} \sqrt{p(w|\boldsymbol{w}_{<t})} \label{eq:theorem1_sub4}\\
&> 2 \log_{2} \sqrt{\prod_{w \in \overline{\mathcal{W}}} (\frac{1}{p(w|\boldsymbol{w}_{<t})})^{p(w|\boldsymbol{w}_{<t})}}
\label{eq:theorem1_sub5}\\
&= \sum_{w \in \overline{\mathcal{W}}} p(w|\boldsymbol{w}_{<t}) \log_{2} p(w|\boldsymbol{w}_{<t}) = \mathrm{H}_{\alpha}|_{\alpha = 1}
\label{eq:theorem1_sub6},
\end{align}
where from Eq.~\eqref{eq:theorem1_sub4} to Eq.~\eqref{eq:theorem1_sub5} we use AM-GM inequality.

\begin{theorem}
\label{theorem:2}
Rényi entropy with $\alpha = 0$ is equivalent to temperature-scaled surprisal with $T \rightarrow \infty$.
\begin{align}
\mathrm{H}_{\alpha}(w_{t}& \mid \boldsymbol{w}_{<t}) |_{\alpha = 0}
= \lim_{T \to \infty} s_{T}(w_{t}, T).
\end{align}
\end{theorem}
\textit{Proof.} By plugging in $\alpha = 0$, Contextual Rényi entropy recovers to be the entropy that readers concentrate on the count of potential words with nonzero probabilities, which is defined in Eq. (5) in~\citet{pimentel-etal-2023-effect}. 
As $T \to \infty$, temperature-scaled surprisal converges to the surprisal induced by random guessing. 
Given the assumtion that $p(w|\boldsymbol{w}_{<t})>0$ for each word $w\in\overline{\mathcal{W}}$, LHS becomes: 
\begin{align}
LHS = -\mathrm{log}_{2}(1/K),
\end{align}
where $K$ is the number of classes. As $T \to \infty$, RHS becomes: 
\begin{align}
RHS &= -\lim_{T \to \infty}
\mathrm{log}_{2}
{\frac{e^{z_{w_{t}}/T}}{\sum_{w \in \mathcal{\overline{\mathcal{W}}}} e^{z_{w}/T}}
}\\
&= - \mathrm{log}_{2}(1/K)
\end{align}

\begin{theorem}
\label{theorem:3}
For $K \geq 2$, the expectation of the $L1$ norm between Rényi entropy with $\alpha = 1$ and temperature-scaled surprisal with $T = 1$ has an upper bound.
\begin{align}
\mathbb{E}[|s_{T}|_{T=1}- \mathrm{H}_{\alpha}|_{\alpha=1}|]
<
\sqrt{\frac{1}{4}\log^{2}(K-1)+1}
\end{align}
\end{theorem}
\textit{Proof.} 
With Jensen's inequality, we have: 
\begin{align}
\mathbb{E}&[|s_{T}|_{T=1}- H_{\alpha}|_{\alpha=1}|] \\
&\leq
\sqrt{\mathbb{E}{[(s_{T}|_{T=1}- H_{\alpha}|_{\alpha=1})^{2}]}} \\
&= 
\sqrt{\mathbb{E}{[
(-\log_{2}p_{w_{t}} - \sum_{w \in \overline{\mathcal{W}}} p(w) (- \log_{2}p(w)))^{2}
]}} \\
&= 
\sqrt{\mathrm{Var}[s_{T}|_{T=1}]} \\
&<
\sqrt{\frac{1}{4}\log^{2}(K-1)+1},
\end{align}
where $\mathrm{Var}[\cdot]$ denotes the variance. The last inequality is shown by Lemma~\ref{lemma:4}, completing the proof of this theorem.

\begin{lemma}\label{lemma:4}
\emph{(Maximum variance of the surprisal).
(See Theorem 8 and Lemma 15 in~\citep{reeb2015tight}).} 
Let $\rho = \mathrm{diag}(p_{1}, p_{2}, . . . , p_{d})$ be a state on a $d$-dimensional system. 
Let $ -\log p_{i}$ be the surprisal of the output $i$ in this system. 
Define $N_{d}$ to be:
\begin{align}
N_{d} := \frac{1}{4} \log^{2}(d-1) + 1.
\end{align}
For $d\geq2$, the variance of surprisal has a tight upper bound: 
\begin{align}
\mathrm{var}_{\rho}(-\log \rho) < N_{d}
\end{align}
\end{lemma}

\begin{figure}[]
\centering 
\hspace{-5mm}
\vspace{-5mm}
\subfigure{
\includegraphics[width=0.7\linewidth]{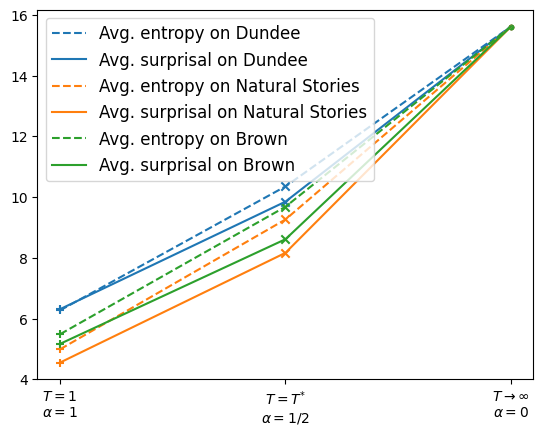}}
\hspace{-5mm}
\caption{A comparison of averaged temperature-scaled surprisal $\overline{s}_{T}|_{T = \{ 1, T^{*}, \infty\}}$ and R\'enyi entropy $\overline{\mathrm{H}}_{\alpha}|_{\alpha = \{0, 1/2, 1 \}}$.}
\label{fig:surprisal_entropy_compare}
\end{figure}

\noindent Theorem~\ref{theorem:2} claims the equivalence of temperature-scaled surprisal $s_{T}(w_{t}, T)$ and Rényi entropy $\mathrm{H}_{\alpha}$ when $T \to \infty$ and $\alpha = 0$. 
Theorem~\ref{theorem:3}, on the other side, gives an upper bound when $T = 1$ and $\alpha = 1$. 
Intuitively, when $T \in (1, \infty)$, $s_{T}$ can be considered as a softened version of $s_{T}|_{T = 1}$. 
Similarly, when $\alpha \in (0, 1)$, $\mathrm{H}_{\alpha}$ can be considered as a softened version of $\mathrm{H}_{\alpha}|_{\alpha = 1}$.
Mathematically, Theorem~\ref{theorem:1} provides the monotonicity of both functions within their respective domains. 
Hypothetically, given the above conditions, when tuning both functions with the aim of a better fit to RTs, $s_{T}|_{T = T^{*}}$ and $\mathrm{H}_{\alpha}|_{\alpha = 1/2}$ might be close. 
Empirically, Fig.~\ref{fig:surprisal_entropy_compare} illustrates the relationship between averaged Rényi entropy $\overline{\mathrm{H}}_{\alpha}|_{\alpha = \{0, 1/2, 1 \}}$ and $\overline{s}_{T}|_{T = \{ 1, T^{*}, \infty\}}$ on probabilities on three corpora.
Notably, $\overline{\mathrm{H}}_{\alpha}|_{\alpha = 1/2}$ and $\overline{s}_{T}|_{T = T^{*}}$ are closely aligned, especially when compared with other entropy and surprisal data points. 
This empirical evidence partly verifies Theorem~\ref{theorem:2}, Theorem~\ref{theorem:3} and our hypothesis.  


\section{Further analysis in Section~\ref{sec:word-level_properties}}
\label{sec:appx_word-level_properties}
We observe that \textbf{larger LMs exhibit an increased $\boldsymbol{\Delta_{\mathrm{MSE}}}$} by utilizing temperature-scaled surprisal, as shown in the average column (Avg.) of Table~\ref{tab:word-level_properties1}. 
Specifically, on Dundee, the top 2 models achieving the largest improvement through temperature scaling are GPT-2 medium and xl, while GPT-2 large and xl have the most benefit on Natural Stories and Brown. 
This result is consistent with previously observed $\Delta_{\mathrm{llh}}$ improvement ($\Delta_{\mathrm{llh}} +$) across the corpus-by-LM reported in Table~\ref{tab:llm_improvement}, \textbf{suggesting a correlation between model likelihood and MSEs of the regression models.} 
We do not observe a mismatch between them, as posited by~\citet{oh2023does} that LME models achieve similar MSEs irrespective of obvious differences in model likelihood.  

Regarding the effect of the change (increase or decrease) of actual-word probability on the final fit to RTs, we first analyzed the ratio of probabilities decreasing (or increasing) for all words, as well as for subsets with specific word-level properties, choosing named entities as the representative, as shown in Table~\ref{tab:word-level_properties2}. 
We observed that \textbf{probabilities of the majority of words (around 80-90\%) decrease by temperature scaling}. 
Compared with the average across all word types (as indicated in the 'Avg.' column), named entities exhibit a lower ratio of probability reduction. 
Larger LMs tend to have a higher ratio, especially the ratio for named entities, likely because smaller models may lack the specific knowledge of less common terms, such as named entities. 

\begin{figure*}[]
\centering 
\subfigure{
\includegraphics[width=0.6\textwidth]{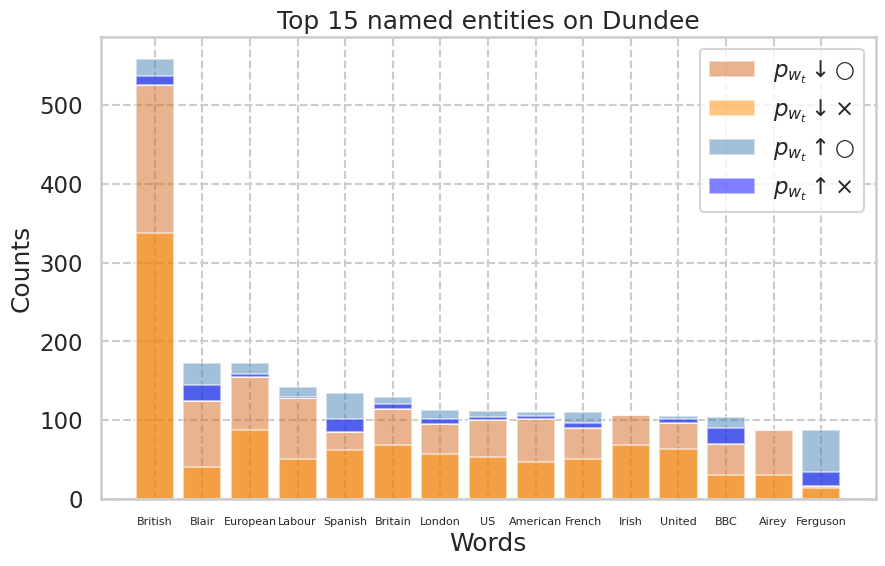}}
\subfigure{
\includegraphics[width=0.6\textwidth]{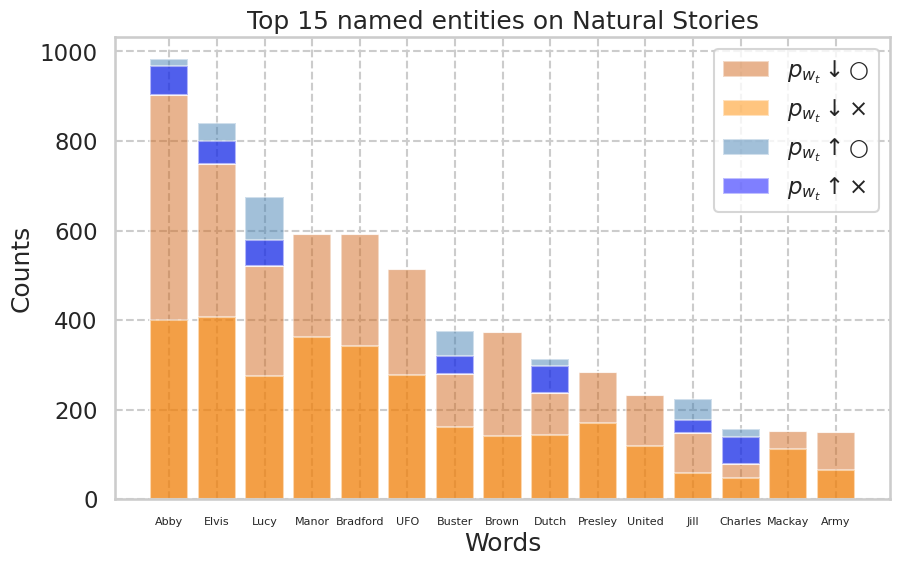}}
\subfigure{
\includegraphics[width=0.6\textwidth]{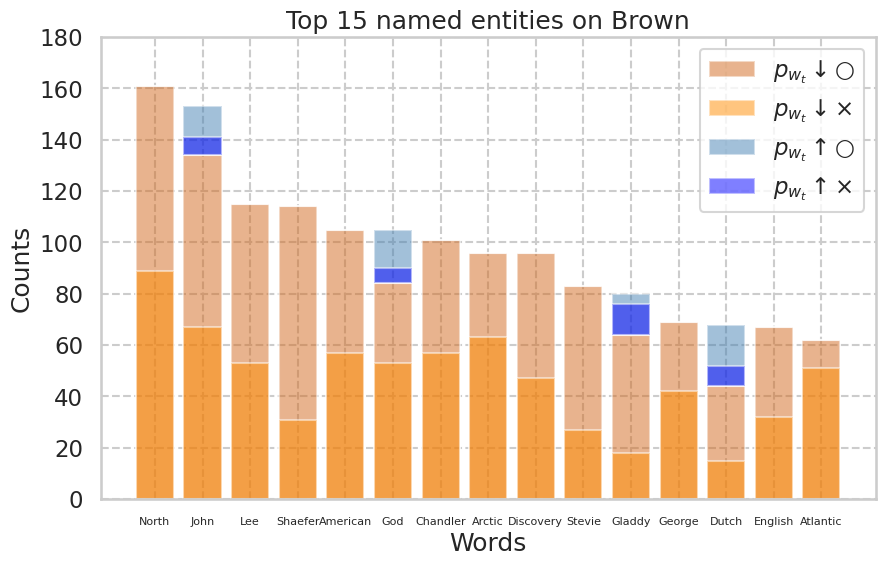}}\\
\caption{Top 15 frequent named entities for GPT-2 small on Dundee, Natural Stories and Brown. $\uparrow$ and $\downarrow$ denote probability being higher and smaller, respectively. $\bigcirc$ and $\times$ denote unbeneficial words (absolute residual error increases) and beneficial words (absolute residual error decreases) by temperature scaling, respectively. }
\label{fig:analysis_top15_words}
\end{figure*}

Recalling one of the results in Section~\ref{sec:word-level_properties} that the main advantage of temperature-scaled surprisal arises from reduction of large probabilities on Dundee and the amplification of small probabilities on Natural Stories. 
However, for named entities, the story is converse on Dundee vs. on Natural Stories and Brown, where for the latter two corpora, the advantage is primarily due to reducing the probabilities of highly predictable entities. 
We shed light to the possible reason of such a discrepancy in Fig.~\ref{fig:analysis_top15_words}, which displays the top 15 frequent words for GPT-2 small on three corpora.
Notably, \textbf{Natural Stories and Brown show a marked lack of words with increased probabilities} (blue bins) compared to Dundee. 
This lack weakens the overall impact of rising probabilities (quantified by $\Delta_{\mathrm{MSE}}(p_{w_{t}} {\uparrow})$). 
Specifically, on Brown, only 4 out of 15 top frequent words have the part of increased probabilities (blue bins), correlating with the largest discrepancy in $\Delta_{\mathrm{MSE}}$ between probabilities that decrease (329.7) and those that increase (-170.6) in Table~\ref{tab:word-level_properties3}.

\begin{table}[h]
\setlength{\tabcolsep}{4pt} 
\centering
\begin{tabular}{clllll}
\hline
\multicolumn{1}{l}{}             &      & \multicolumn{2}{c}{Avg.}         & \multicolumn{2}{c}{Named entities} \\
\multicolumn{1}{c}{Corpus}             & GPT2 & $p_{w_t}$  $\downarrow$ & $|res|$ $\downarrow$ & $p_{w_t}$ $\downarrow$  & $|res|$ $\downarrow$ \\ \hline
\multirow{4}{*}{Dundee}          & s    & 88.0          & 51.8             & 78.1            & {52.3}    \\
 & m    & 89.6          & 52.5             & 80.1            & {54.1}    \\
 & l    & 90.2          & 52.3             & 80.1            & {53.5}    \\
 & xl   & 91.4          & 52.4             & 82.7            & {54.3}    
 \\ \hline
 & s    & 93.8          & {55.0}    & 85.3            & 51.8             \\
Natural & m    & 94.7          & {55.2}    & 89.1            & 53.2             \\
Stories & l    & 93.5          & {55.7}    & 89.1            & 53.4             \\
 & xl   & 92.1          & {55.5}    & 88.2            & 52.8             
 \\ \hline
\multirow{4}{*}{Brown}           & s    & 91.8          & {51.5}    & 87.3            & 50.9             \\
 & m    & 93.2          & {51.5}    & 86.1            & 50.9             \\
 & l    & 93.3          & {51.8}    & 88.6            & 52.1             \\
 & xl   & 93.5          & {51.7}    & 87.8            & 53.3             
 \\ \hline
\end{tabular}
\caption{The ratio of probability of predicted word $p_{w_t}$ getting smaller and the absolute value of residuals $|res|$ getting smaller for GPT-2 models on three corpora. 
}
\label{tab:word-level_properties2}
\end{table}

\section{Preprocessing steps}
\label{sec:appx1}
On Dundee ET corpus~\citep{kennedy2003dundee}, we use the first-pass gaze duration. 
Following prior work~\citep{kuribayashi2022context}, we remove words containing numbers or punctuation, words that are either the first or the last one in a line, as well as words whose previous words contain numbers or punctuation. 
On Natural Stories SPR corpus~\citep{futrell2018natural}, following~\citet{shain2022large}, we remove words if the RT is less than 100ms or greater than 3,000ms, if the words are in the first or last position of each story, if participants answered less than 5 out of 8 comprehension questions correctly, if words contain numbers or punctuation, and if words whose previous words containing numbers or punctuation. 
On Brown SPR corpus~\citep{smith2013effect}, following~\citet{shain2022large}, we remove words if the RT is less than 100ms or greater than 3,000ms and if words contain numbers or punctuation.

\section{Further descriptions on metrics and evaluation}
\label{sec:appx_other_metric}
We evaluate calibration error (\% ECE and \% CECE) in both equally-spaced and log binning schemes. 
In equally-spaced binning scheme, the samples are grouped into $M \in \mathbb{N}$ equally-spaced interval bins based on their confidences $\hat{p}_{i}$. 
Conversely, the log binning scheme operates under an \textit{empirical upper limit} for $-\log_{2}{\hat{p}_{i}}$, denoted as $\max(-\log_{2}{\hat{\boldsymbol{p}}})$. 
Table~\ref{tab:p_ranges} shows ranges of $\hat{\boldsymbol{p}}$ and $-\log_{2}{\hat{\boldsymbol{p}}}$ for GPT2s on three corpora. 
For this scheme, we establish $M \in \mathbb{N}$ log-equally-spaced interval bins within the range of (0, $\max(-\log_{2}{\hat{\boldsymbol{p}}})$]. 

We investigate scaling $T \in [1, 10]$, considering both densely and sparsely distributed points. The values examined are detailed as follows: [1.0, 1.1, ..., 1.9] for dense intervals, [2.0, 2.25, ..., 3.25] for moderately spaced intervals, and [3.5, 4.0, ..., 10.0] for sparse intervals.

Following~\citet{kuribayashi2022context}, reading times of a base model are modelled by the following formula:
\begin{align}
\begin{split}
    rt &\sim \text{freq} \ast \text{length} + \text{freq\_prev\_1} \ast \text{length\_prev\_1} \\
    & + (1|\text{article}) + (1|\text{subj\_id})
\end{split}
\end{align}
A target model additionally includes surprisal estimates of current words and previous words: 
\begin{align}
\begin{split}
    rt &\sim \text{surprisal} + \text{surprisal\_prev\_1} + \text{surprisal\_prev\_2} \\
    & + \text{freq} \ast \text{length} + \text{freq\_prev\_1} \ast \text{length\_prev\_1} \\
    & + (1|\text{article}) + (1|\text{subj\_id}).
\end{split}
\label{eq:model1}
\end{align}
On Dundee corpus, both models also include features of $[\text{screenN}, \text{lineN}, \text{segmentN}]$. 
We also perform experiments with both models without interactions among predictors in Appendix~\ref{sec:appx_interactions_correlations_predictors}. 

\begin{table*}[]
\centering
\begin{tabular}{ccc}
\hline
& $\hat{\boldsymbol{p}}$     & $-\log_{2}{\hat{\boldsymbol{p}}}$ \\ \hline
Dundee & [4.99e-03, 1) & (0, 7.65]  
\\\hline
Natural Stories & [8.567e-03, 1) & (0, 6.87]
\\\hline
Brown &[8.15e-03, 1) & (0, 6.94] \\\hline
\end{tabular}
\caption{Ranges of $\hat{p}$ and $-\log_{2}{\hat{p}}$ for GPT2s on Dundee, Natural Stories and Brown.}
\label{tab:p_ranges}
\end{table*}

\section{Exploring further effectiveness of temperature-scaled surprisal over basic predictors}

\begin{figure*}[]
\centering 
\hspace{-5mm}
\subfigure{
\includegraphics[width=0.35\textwidth]{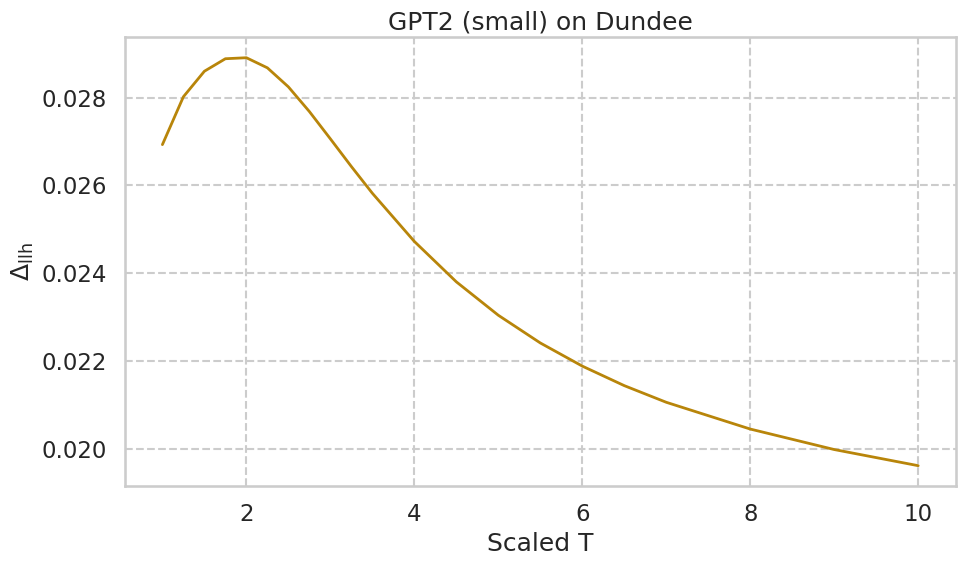}}
\hspace{-5mm}
\subfigure{
\includegraphics[width=0.35\textwidth]{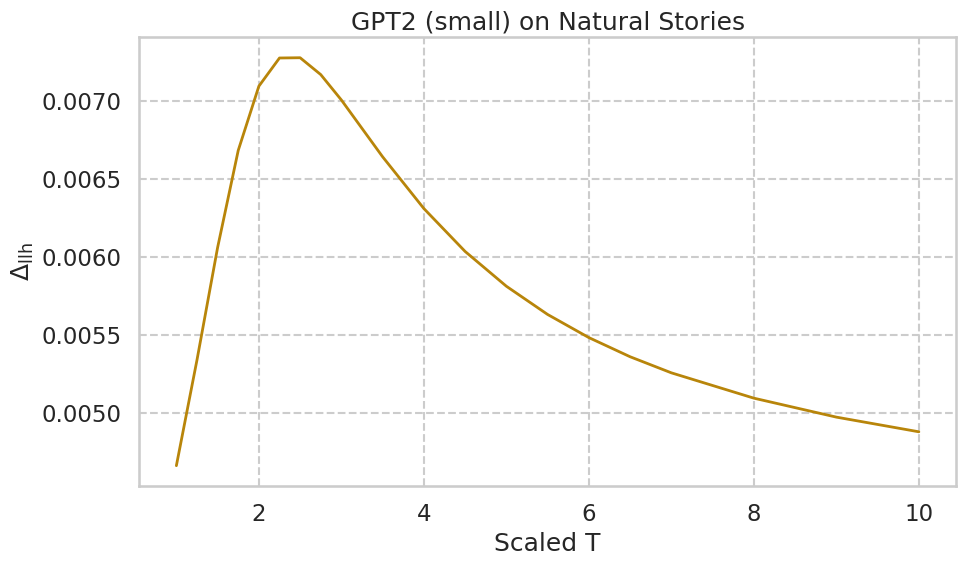}}
\hspace{-5mm}
\subfigure{
\includegraphics[width=0.35\textwidth]{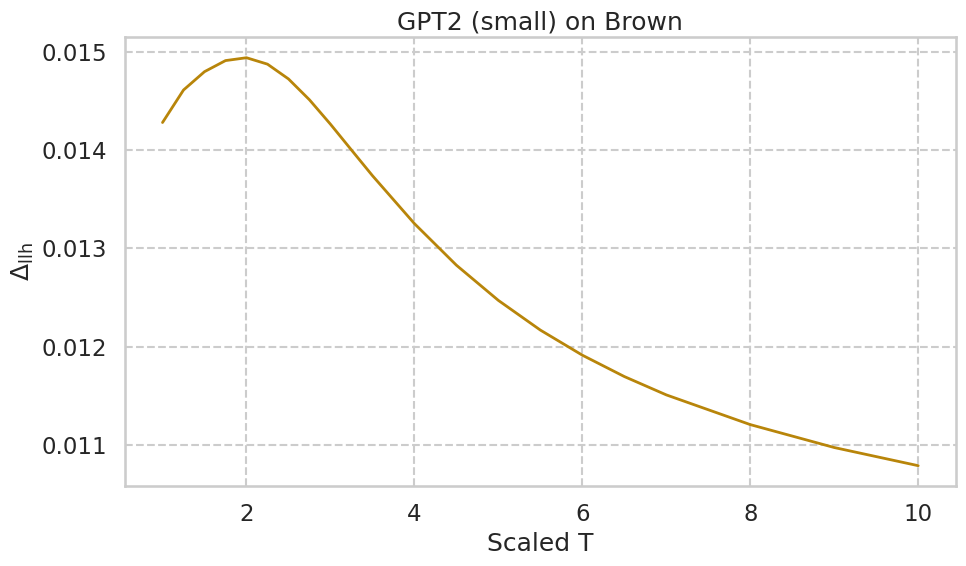}}
\hspace{-5mm}
\vspace{-3mm}
\caption{Relationship between $\Delta_{\mathrm{llh}}$ of GPT-2 small and corresponding temperature. T is scaled from 1.0 to 10. Base predictor variables $\mathbf{v}^{base}$ and target predictor variables are 0 and temperature-scaled surprisal $s_{T}(w_{t}, T)$, respectively. }
\label{fig:appendix_predictor0}
\end{figure*}

In this section, we explore the question of whether the benefit of temperature-scaled surprisal holds only for regression models already containing other predictors such as length and frequency. 
We conduct experiments similar to those detailed in Section~\ref{sec:results1} while setting base predictor variables $\mathbf{v}^{base}$ to 0 and target predictor variables $\mathbf{v}^{tgt}$ to only temperature-scaled surprisal $s_{T}(w_{t}, T)$ in Eq.~\ref{eq:PPP}. 
Fig.~\ref{fig:appendix_predictor0} shows that \textbf{excluding base predictors decrease but not totally impact the effectiveness of temperature-scaled surprisal.}

\section{Calibration error for single-token and multiple-token words}

In Table~\ref{tab:cal_table2}, we demonstrate the calibration error (\%ECE) for single-token and multiple-token words for GPT-2 small. 
Calibration evaluation is conducted at the token level as before. 
Results indicate that \textbf{multiple-token words show larger calibration errors than single-token words. }

\begin{table*}[]
\centering
\begin{tabular}{llll}
\hline
&      & $\mathrm{ECE}_{\mathrm{single}}$ & $\mathrm{ECE}_{\mathrm{multiple}}$ \\ \hline
\multirow{2}{*}{Dundee} & $T=1$  & 1.98        & 2.05          \\
& $T=T*$ & 25.58       & 36.10         \\ \hline
\multicolumn{1}{c}{\multirow{2}{*}{Natural Stories}} & $T=1$  & 2.20        & 3.78          \\
\multicolumn{1}{c}{}                                 & $T=T*$ & 32.38       & 47.02         \\ \hline
\multirow{2}{*}{Brown}                               & $T=1$  & 1.69        & 3.86          \\
& $T=T*$ & 28.70       & 42.99         \\ \hline
\end{tabular}
\caption{Expected calibration errors of tokens in single-token (\% $\mathrm{ECE}_{\mathrm{single}}$) and multiple-token words (\% $\mathrm{ECE}_{\mathrm{multiple}}$) before and after temperature scaling for GPT-2 small on Dundee, Natural Stories and Brown. Results are all evaluated on the equally-spaced binning scheme. }
\label{tab:cal_table2}
\end{table*}

\section{Probability distribution before and after temperature scaling}

\begin{figure*}[]
\centering 
\subfigure{
\includegraphics[width=1.0\textwidth]{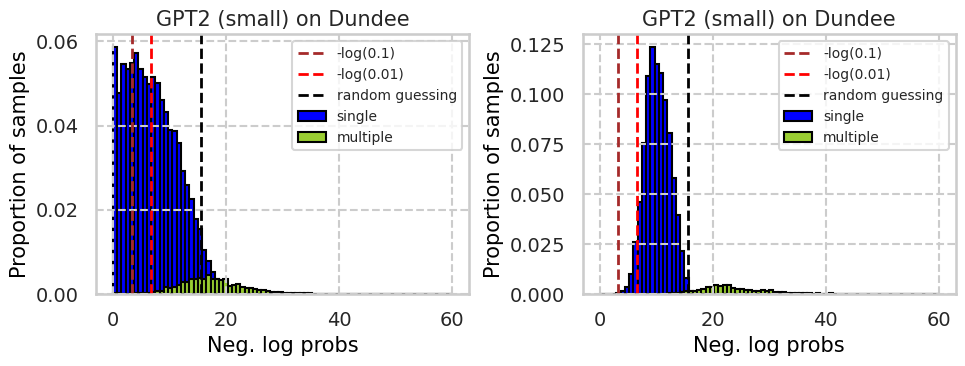}}
\subfigure{
\includegraphics[width=1.0\textwidth]{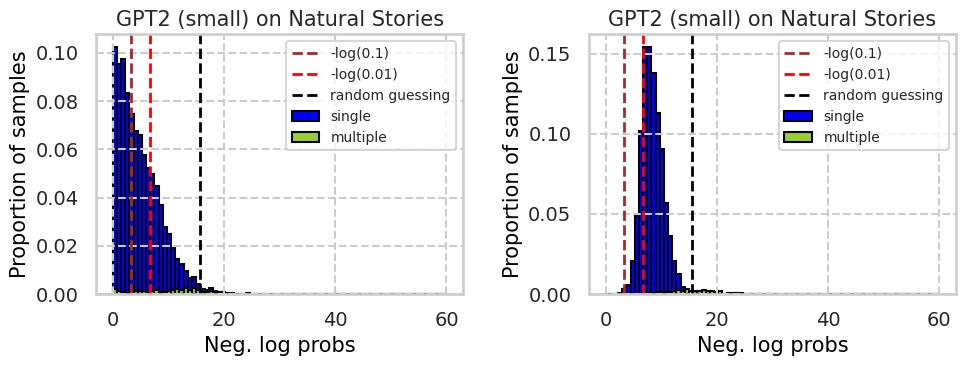}}
\subfigure{
\includegraphics[width=1.0\textwidth]{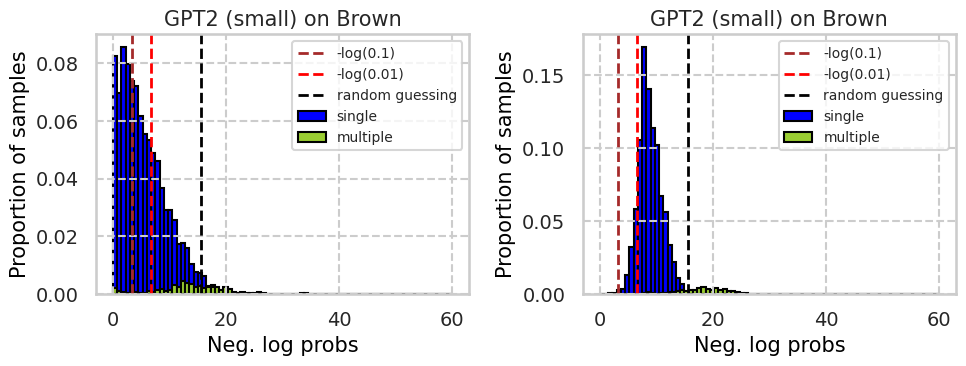}}
\caption{Distribution of negative log actual-word probability (surprisal) before (left side of figure) and after (right side of figure) temperature scaling for single-token and multiple-token words for GPT-2 small on three corpora. Values of surprisal with probability of 0.1, 0.01 and 1/K (random guessing) are displayed using dash lines. }
\label{fig:prob_dis}
\end{figure*}

Fig.~\ref{fig:prob_dis} shows actual-word probability distribution before and after temperature scaling for GPT-2 small on three corpora. 
\textbf{Multiple-token words tend to have smaller probabilities than single-token words}, both before and after temperature scaling. 

\section{Significant test of temperature-scaled surprisal}
\label{sec:appx_significance}
We report the statistical significance based on selecting the most representative model, GPT2s, on three corpora in Table~\ref{tab:p_values}. Models with temperature-scaled surprisal lead to statistically significant positive $\Delta_{\mathrm{llh}}$ (p < 0.001). 

\begin{table*}[]
\centering
\begin{tabular}{lcc}
\hline
Corpora & Models          & p               \\ \hline
Dundee  & target vs. base & \textless 0.001 \\
NS      & target vs. base & \textless 0.001 \\
Brown   & target vs. base & \textless 0.001 \\ \hline
\end{tabular}
\caption{Significance of temperature-scaled surprisal for GPT2 small on three corpora with $T = T^{*}$. }
\label{tab:p_values}
\end{table*}


\section{Analysis on correlations among predictors}
\label{sec:appx_interactions_correlations_predictors}
We investigate the question of whether the benefit of temperature-scaled surprisal is primarily due to the interactions and correlations among predictors. 
We first run experiments with the original target LME model as in Eq.~\ref{eq:model1} (denoted as \textit{model 1}), a model that has no interactions between frequency and length as in Eq.~\ref{eq:model2} (denoted as \textit{model 2}) and a third model that has no interactions and additionally includes random slopes for subject as in Eq.~\ref{eq:model3} (denoted as \textit{model 3}). 

\begin{align}
\begin{split}
    rt &\sim \text{surprisal} + \text{surprisal\_prev\_1} + \text{surprisal\_prev\_2} \\
    & + \text{freq} + \text{length} + \text{freq\_prev\_1} + \text{length\_prev\_1} \\
    & + (1|\text{article}) + (1|\text{subj\_id}).
\end{split}
\label{eq:model2}
\end{align}

\begin{align}
\begin{split}
    rt &\sim \text{surprisal} + \text{surprisal\_prev\_1} + \text{surprisal\_prev\_2} \\
    & + \text{freq} + \text{length} + \text{freq\_prev\_1} + \text{length\_prev\_1} \\
    & + (1|\text{article}) + (\text{surprisal}|\text{subj\_id}).
\end{split}
\label{eq:model3}
\end{align}

The results are in Table~\ref{tab:interations_lme}. \textbf{Removing the interactions among predictors or additionally including random slopes does not influence the effectiveness of temperature-scaled surprisal. }

\begin{table*}[]
\centering
\begin{tabular}{lllccr}
\hline
Corpora & Models & $T^{*}$ & $\Delta_{\mathrm{llh}}(T = 1)$ & $\Delta_{\mathrm{llh}}(T = T^{*})$ & $\Delta_{\mathrm{llh}}+$ \\ \hline
Dundee  & model1 & 2.75 & 6.90                 & 8.45                 & 22.5 \\
Dundee  & model2 & 2.75 & 6.79                 & 8.12                 & 19.6 \\
Dundee  & model3 & 2.75 & 7.81                 & 9.12                 & 16.8 \\ \hline
Natural Stories & model1 & 2.5  & 4.36                 & 6.99                 & 60.3 \\
Natural Stories & model2 & 2.5  & 4.35                 & 6.99                 & 60.7 \\
Natural Stories & model3 & *  & *                    & *                    & *    \\ \hline
Brown   & model1 & 2.5  & 6.62                 & 7.53                 & 13.7 \\
Brown   & model2 & 2.25  & 6.62                 & 7.30                 & 10.3  \\
Brown   & model3 & *  & *                    & *                    & *    \\ \hline
\end{tabular}
\caption{Optimal $T^{*}$, $\Delta_{\mathrm{llh}}(T = 1)$, $\Delta_{\mathrm{llh}}(T = T^{*})$, and $\Delta_{\mathrm{llh}}+$ for three models for GPT2 small on three corpora. $*$ indicates regression models not converged. }
\label{tab:interations_lme}
\end{table*}

Furthermore, we also investigated the correlations among predictors by examining the correlation matrix for GPT2 small on three corpora (model 1). Table~\ref{tab:interations_lme_Dundee}, \ref{tab:interations_lme_NS} and \ref{tab:interations_lme_Brown} indicate that \textbf{temperature-scaled surprisal does not exhibit a stronger correlation with the other predictors in comparison to the original surprisal}, as shown in the surprisal column ('surp'), which excludes the concern that the primary benefits are simply due to correlations between the baseline predictor and temperature-scaled surprisal.

\begin{figure*}
\centering
\subfigure[$T = 1$]{
\begin{tabular}{llllllllll}
\hline & (Intr) & surp   & surp\_1 & surp\_2 & log\_frq & length & log\_frq\_1 & length\_1 & log\_frq\_2 \\ \hline
surp        & 0.004  &        &         &         &          &        &             &           &             \\
surp\_1     & 0.000  & -0.147 &         &         &          &        &             &           &             \\
surp\_2     & -0.001 & -0.057 & -0.101  &         &          &        &             &           &             \\
log\_frq    & 0.0200   & 0.238  & 0.002   & -0.03   &          &        &             &           &             \\
length      & 0.019  & -0.272 & 0.027   & 0.04    & 0.602    &        &             &           &             \\
log\_frq\_1 & 0.022  & -0.085 & 0.332   & -0.048  & 0.034    & -0.021 &             &           &             \\
length\_1   & 0.028  & 0.034  & -0.200    & 0.031   & 0.003    & -0.025 & 0.650        &           &             \\
log\_frq\_2 & 0.032  & -0.081 & 0.002   & 0.000   & 0.374    & 0.626  & -0.009      & 0.014     &             \\
length\_2   & 0.038  & -0.013 & -0.033  & 0.003   & -0.003   & 0.043  & 0.509       & 0.578     & 0.014       \\ \hline
\end{tabular}
}
\subfigure[$T = T^{*}$]{
\begin{tabular}{llllllllll}
\hline & (Intr) & surp   & surp\_1 & surp\_2 & log\_frq & length & log\_frq\_1 & length\_1 & log\_frq\_2 \\ \hline
surp        & 0.006  &        &         &         &          &        &             &           &             \\
surp\_1     & 0.005  & -0.145 &         &         &          &        &             &           &             \\
surp\_2     & -0.003 & -0.074 & -0.154  &         &          &        &             &           &             \\
log\_frq    & 0.020   & -0.055 & 0.050    & -0.013  &          &        &             &           &             \\
length      & 0.017  & -0.395 & 0.042   & 0.044   & 0.676    &        &             &           &             \\
log\_frq\_1 & 0.024  & -0.058 & 0.063   & 0.011   & 0.051    & -0.018 &             &           &             \\
length\_1   & 0.025  & 0.060   & -0.353  & 0.075   & -0.016   & -0.035 & 0.702       &           &             \\
log\_frq\_2 & 0.031  & -0.156 & 0.004   & 0.004   & 0.409    & 0.634  & -0.005      & 0.011     &             \\
length\_2   & 0.037  & 0.001  & -0.088  & -0.006  & -0.003   & 0.038  & 0.542       & 0.574     & 0.014       \\ \hline
\end{tabular}
}
\caption{Correlation matrix for GPT2s on Dundee with (a) $T = 1$ and (b) $T = T^{*}$. }
\label{tab:interations_lme_Dundee}
\end{figure*}

\begin{figure*}
\centering
\subfigure[$T = 1$]{
\begin{tabular}{llllllllll}
\hline & (Intr) & surp   & surp\_1 & surp\_2 & log\_frq & length & log\_frq\_1 & length\_1 & log\_frq\_2 \\ \hline
surp        & 0.002  &        &         &         &          &        &             &           &             \\
surp\_1     & 0.002  & -0.009 &         &         &          &        &             &           &             \\
surp\_2     & 0.001  & 0.003  & -0.019  &         &          &        &             &           &             \\
log\_frq    & 0.017  & 0.237  & 0.013   & -0.016  &          &        &             &           &             \\
length      & 0.022  & -0.181 & 0.018   & 0.011   & 0.692    &        &             &           &             \\
log\_frq\_1 & 0.018  & 0.019  & 0.238   & -0.051  & 0.067    & -0.015 &             &           &             \\
length\_1   & 0.022  & 0.013  & -0.183  & 0.029   & -0.01    & 0.011  & 0.672       &           &             \\
log\_frq\_2 & 0.030   & 0.005  & 0.030    & 0.010    & 0.472    & 0.586  & 0.008       & 0.017     &             \\
length\_2   & 0.030   & 0.010   & 0.011   & 0.018   & -0.005   & 0.02   & 0.468       & 0.589     & 0.023       \\ \hline
\end{tabular}
}
\subfigure[$T = T^{*}$]{
\begin{tabular}{llllllllll}
\hline
            & (Intr) & surp   & surp\_1 & surp\_2 & log\_frq & length & log\_frq\_1 & length\_1 & log\_frq\_2 \\ \hline
surp        & 0.013  &        &         &         &          &        &             &           &             \\
surp\_1     & 0.011  & -0.108 &         &         &          &        &             &           &             \\
surp\_2     & -0.002 & -0.034 & -0.080   &         &          &        &             &           &             \\
log\_frq    & 0.020   & 0.200    & 0.009   & -0.021  &          &        &             &           &             \\
length      & 0.020   & -0.194 & 0.014   & 0.010    & 0.700      &        &             &           &             \\
log\_frq\_1 & 0.019  & -0.09  & 0.231   & -0.026  & 0.048    & 0.001  &             &           &             \\
length\_1   & 0.020   & 0.016  & -0.203  & 0.045   & -0.013   & 0.014  & 0.667       &           &             \\
log\_frq\_2 & 0.031  & 0.035  & 0.004   & -0.007  & 0.482    & 0.578  & 0.000           & 0.020      &             \\
length\_2   & 0.031  & 0.015  & 0.038   & -0.036  & -0.003   & 0.019  & 0.474       & 0.579     & 0.023       \\ \hline
\end{tabular}
}
\caption{Correlation matrix for GPT2s on Natural Stories with (a) $T = 1$ and (b) $T = T^{*}$. }
\label{tab:interations_lme_NS}
\end{figure*}

\begin{figure*}
\centering
\subfigure[$T = 1$]{
\begin{tabular}{llllllllll}
\hline
            & (Intr) & surp   & surp\_1 & surp\_2 & log\_frq & length & log\_frq\_1 & length\_1 & log\_frq\_2 \\ \hline
surp        & 0.003  &        &         &         &          &        &             &           &             \\
surp\_1     & -0.003 & -0.058 &         &         &          &        &             &           &             \\
surp\_2     & -0.001 & -0.021 & -0.039  &         &          &        &             &           &             \\
log\_frq    & 0.032  & 0.269  & 0.007   & -0.053  &          &        &             &           &             \\
length      & 0.036  & -0.206 & 0.005   & -0.007  & 0.691    &        &             &           &             \\
log\_frq\_1 & 0.007  & -0.068 & 0.212   & -0.044  & 0.084    & 0.021  &             &           &             \\
length\_1   & 0.012  & -0.018 & -0.379  & 0.036   & 0.022    & 0.060   & 0.484       &           &             \\
log\_frq\_2 & 0.045  & -0.003 & 0.000   & -0.009  & 0.539    & 0.593  & -0.013      & 0.016     &             \\
length\_2   & 0.028  & -0.019 & -0.09   & 0.018   & 0.020     & 0.054  & 0.247       & 0.347     & -0.012      \\ \hline
\end{tabular}
}
\subfigure[$T = T^{*}$]{
\begin{tabular}{llllllllll}
\hline
            & (Intr) & surp   & surp\_1 & surp\_2 & log\_frq & length & log\_frq\_1 & length\_1 & log\_frq\_2 \\ \hline
surp        & 0.019  &        &         &         &          &        &             &           &             \\
surp\_1     & -0.010  & -0.114 &         &         &          &        &             &           &             \\
surp\_2     & -0.002 & -0.046 & -0.096  &         &          &        &             &           &             \\
log\_frq    & 0.035  & 0.165  & 0.010    & -0.049  &          &        &             &           &             \\
length      & 0.032  & -0.241 & 0.027   & -0.010   & 0.719    &        &             &           &             \\
log\_frq\_1 & 0.008  & -0.103 & -0.124  & 0.011   & 0.078    & 0.034  &             &           &             \\
length\_1   & 0.015  & 0.019  & -0.572  & 0.079   & 0.018    & 0.043  & 0.580        &           &             \\
log\_frq\_2 & 0.045  & 0.015  & -0.024  & -0.023  & 0.554    & 0.584  & -0.012      & 0.026     &             \\
length\_2   & 0.029  & 0.009  & -0.263  & 0.008   & 0.022    & 0.046  & 0.295       & 0.418     & -0.005      \\ \hline
\end{tabular}}
\caption{Correlation matrix for GPT2s on Brown with (a) $T = 1$ and (b) $T = T^{*}$. }
\label{tab:interations_lme_Brown}
\end{figure*}

\section{Influence of multiple-token words vs.~model size}
\label{appx:mult-tok-model-size}
\begin{table}[]
\tabcolsep=2pt
\centering
\small
\begin{tabular}{ccc}
\hline
\multicolumn{1}{l}{}             & GPT2 & $\Delta_{\mathbf{llh}}+(\mathrm{multiple})$ \\ \hline
\multirow{4}{*}{Dundee}          & s  & 23.6 \\
                                 & m  & 36.4 \\
                                 & l  & 38.0 \\
                                 & xl & 42.9 \\ \hline
\multirow{4}{*}{NS}              & s  & 45.2 \\
                                 & m  & 50.1 \\
                                 & l  & 62.0 \\
                                 & xl & 67.8 \\ \hline
\multirow{4}{*}{Brown}           & s  & 9.2 \\
                                 & m  & 13.4 \\
                                 & l  & 17.9 \\
                                 & xl & 5.49 \\ \hline
\end{tabular}
\caption{$\Delta_{\mathrm{llh}}$ improvement by only scaling tokens in multiple-token words (\%) ($\Delta_{\mathrm{llh}} + (\mathrm{multiple}) = (\Delta_{\mathrm{llh}}(T=T^{*}, \mathrm{multiple}) - \Delta_{\mathrm{llh}}(T=1))/\Delta_{\mathrm{llh}}(T=1)$) for GPT2s on Dundee, Natural Stories (NS) and Brown.}
\label{tab:llm_improvement_multiple}
\end{table}

Table~\ref{tab:llm_improvement_multiple} shows the increase of $\Delta_{\mathrm{llh}}$ of temperature-scaled surprisal by only taking into the analysis the subset of multiple-token words. 
The benefit of temperature-scaled surprisal being primarily from the scaling of multiple-token words still holds for larger LLMs. 
\textbf{For larger LLMs, the influence of multiple-token words is also larger.}

\section{Influence of word-level attributes vs.~influence of multiple-token words}
\label{appx:ne-vs-multitok}
We explore which of these two factors has a stronger effect on the benefit of temperature-scaled surprisal, word-level attributes in Section~\ref{sec:word-level_properties} or multiple-token words in Section~\ref{sec:multiple-token_words}. 
For word types, we select named entities as the representative attribute since they perform to be the most beneficial ones as discussed in Section~\ref{sec:word-level_properties}. 
For multiple-token words, we select all multiple-token words with more-than-one tokens. 
In order to fairly compare the influence, we normalize $\Delta_{\mathrm{MSE}}$ of each category under the linguistic factor $F$ with the ratio of that category words among the total words: $\Bar{\Delta}_{\mathrm{MSE}}(F) = \Delta_{\mathrm{MSE}(F)} \cdot \mathrm{ratio}(F)$. 
Table~\ref{tab:word_level_properties_vs_multiple} shows that 
\textbf{multiple-token words drive the much stronger averaged benefit of temperature-scaled surprisal}, compared with the averaged benefit of named entities. 

\begin{table}[]
\tabcolsep=2pt
\centering
\small
\begin{tabular}{clll}
\hline
\multicolumn{1}{l}{}    & GPT2 & NE   & \#\textgreater 1 \\ \hline
\multirow{4}{*}{Dundee} & s    & 3.9  & 17.0              \\
                        & m    & 6.9  & 26.7              \\
                        & l    & 7.2  & 27.0              \\
                        & xl   & 7.6  & 27.9              \\ \hline
\multirow{4}{*}{NS}     & s    & 2.6  & 35.9              \\
                        & m    & 2.2  & 38.4              \\
                        & l    & 2.1  & 43.3              \\
                        & xl   & 2.0  & 40.6              \\ \hline
\multirow{4}{*}{Brown}  & s    & 10.2 & 27.0              \\
                        & m    & 9.8  & 28.9              \\
                        & l    & 10.1 & 30.7              \\
                        & xl   & 10.8 & 36.0              \\ \hline
\end{tabular}
\caption{$\Bar{\Delta}_{\mathrm{MSE}}$ measurement on named entites (NE) and multiple-token words (\#>1) for GPT-2 models on Dundee, Natural Stories (NS) and Brown. }
\label{tab:word_level_properties_vs_multiple}
\end{table}

\section{Other results in Section~\ref{sec:discussion}}
\label{sec:appx_attached_figs_tables}

\begin{table*}[]
\setlength{\tabcolsep}{3pt} 
\centering
\begin{tabular}{l|llll|llll}
\hline
& \multicolumn{4}{c|}{ratio of $p_{w_{t}} {\downarrow}$} & \multicolumn{4}{c}{ratio of named entities}        \\ \cline{2-9} 
& \#=1  & \#\textgreater{}1 & \#=2 & \#=3 & \#=1 & \#\textgreater{}1 & \#=2 & \#=3 \\ \hline
Dundee          & 87.6          & 93.7              & 90.6   & 98.3   & 3.7          & 16.3              & 16.6   & 17.4   \\
Natural Stories & 92.1          & 93.0              & 92.2   & 97.2*  & 1.3          & 3.5               & 3.3    & 4.7*   \\
Brown           & 93.0          & 98.1              & 97.6   & 35.2*  & 3.3          & 12.3              & 10.9   & 17.0*  \\ \hline
\end{tabular}
\caption{This table displays the ratio of words with decreasing probability ($p_{w_{t}} {\downarrow}$) and the ratio of named entities on subsets for both single-token words (\#=1) and multiple-token words (\#>1) for GPT-2 small on three corpora.
Numbers marked with $*$ indicate subsets with insufficient (less than 1\%) data.}
\label{tab:influence_of_multipletoken_words}
\end{table*}

\begin{table*}[]
\setlength{\tabcolsep}{3pt} 
\centering
\begin{tabular}{lllllllll}
\hline
& \multicolumn{2}{c}{\#=1}                         & \multicolumn{2}{c}{\#\textgreater{}1}            & \multicolumn{2}{c}{\#=2}                         & \multicolumn{2}{c}{\#=3}                         \\
& $p_{w_{t}} {\downarrow}$ & $p_{w_{t}} {\uparrow}$ 
& $p_{w_{t}} {\downarrow}$ & $p_{w_{t}} {\uparrow}$ 
& $p_{w_{t}} {\downarrow}$ & $p_{w_{t}} {\uparrow}$ 
& $p_{w_{t}} {\downarrow}$ & $p_{w_{t}} {\uparrow}$
\\ \hline
Dundee & 8.0 & 19.6 & 269.5 & -20.3* & 50.5 & 26.6* & 497.4 & 125.4** 
\\
NS & 117.3 & 142.3 & 242.5 & 93.0* & 312.6 & 95.8* & -123.9* &   50.6** 
\\
Brown & 35.2 & -61.0 & 327.3 & 5290.2** & 17.3 & 5290.2** & 655.0* & 0.0** 
\\ \hline
\end{tabular}
\caption{
Given words with decreasing (and increasing) probability, the corresponding $\Delta_{\mathrm{MSE}}(p_{w_t} {\downarrow})$ (and $\Delta_{\mathrm{MSE}}(p_{w_t} {\uparrow})$) measurement for both single-token words (\#=1) and multiple-token words (\#>1) for GPT-2 small on three corpora. 
Numbers marked with $*$ indicate subsets with insufficient (less than 1\%) data.
Numbers marked with $**$ indicate subsets with super insufficient (around or less than 0.1\%) data.}
\label{tab:mse_multipletokens}
\end{table*}

\end{document}